\documentclass[nohyperref]{article}

\usepackage{microtype}
\usepackage{graphicx}
\usepackage{subfigure}
\usepackage{booktabs} 
\usepackage{dsfont}

\usepackage{hyperref}

\usepackage[accepted]{icml2022}

\usepackage{amsmath}
\usepackage{amssymb}
\usepackage{mathtools}
\usepackage{amsthm}

\usepackage[capitalize]{cleveref}
\crefname{section}{Section}{Sections} 

\theoremstyle{plain}

\theoremstyle{definition}

\theoremstyle{remark}

\usepackage[textsize=tiny]{todonotes}

\usepackage{color,soul}

\DeclareMathAlphabet\mathbfcal{OMS}{cmsy}{b}{n}

\icmltitlerunning{Short-Term Plasticity Neurons}

\begin{document}

\twocolumn[
\icmltitle{Short-Term Plasticity Neurons Learning to Learn and Forget}




\begin{icmlauthorlist}
\icmlauthor{Hector Garcia Rodriguez}{huazur,ucl}
\icmlauthor{Qinghai Guo}{huashen}
\icmlauthor{Timoleon Moraitis}{huazur}
\end{icmlauthorlist}

\icmlaffiliation{huazur}{Huawei Technologies -- Zurich Research Center, Switzerland}
\icmlaffiliation{huashen}{Advanced Computing \& Storage Lab, Huawei Technologies, Shenzhen, China}
\icmlaffiliation{ucl}{University College London, United Kingdom}

\icmlcorrespondingauthor{Timoleon Moraitis}{timoleon.moraitis@huawei.com}

\icmlkeywords{Neuromorphic Computing, Machine Learning, short-term plasticity, biologically inspired, neuroscience, recurrent neural networks, reinforcement learning, lstm, Hebbian plasticity, AI}

\vskip 0.3in
]



\printAffiliationsAndNotice{}  

\begin{abstract}
Short-term plasticity (STP) is a mechanism that stores decaying memories in synapses of the cerebral cortex. In computing practice, STP has been used, but mostly in the niche of spiking neurons, even though theory predicts that it is the optimal solution to certain dynamic tasks. Here we present a new type of recurrent neural unit, the STP Neuron (STPN), which indeed turns out strikingly powerful. Its key mechanism is that synapses have a state, propagated through time by a self-recurrent connection-within-the-synapse. This formulation enables training the plasticity with backpropagation through time, resulting in a form of learning to learn and forget in the short term. The STPN outperforms all tested alternatives, i.e.\ RNNs, LSTMs, other models with fast weights, and differentiable plasticity. We confirm this in both supervised and reinforcement learning (RL), and in tasks such as Associative Retrieval, Maze Exploration, Atari video games, and MuJoCo robotics. Moreover, we calculate that, in neuromorphic or biological circuits, the STPN minimizes energy consumption across models, as it depresses individual synapses dynamically. Based on these, biological STP may have been a strong evolutionary attractor that maximizes both efficiency and computational power. The STPN now brings these neuromorphic advantages also to a broad spectrum of machine learning practice. 
Code is available at  \url{https://github.com/NeuromorphicComputing/stpn}.
\end{abstract}
~\\
\section{Introduction}
\label{sec:intro}
\subsection{Biological vs artificial neural networks}
Biological neural networks are the source of inspiration for some of the most successful machine learning (ML) models, namely the artificial neural networks (ANNs) that underpin Deep Learning. Despite the success of ANNs, artificial intelligence (AI) models still pale in comparison to animals and humans in several respects. Widely acknowledged limitations include (a) the vast number of training episodes required before mastering a task, (b) difficulties in dynamically changing tasks, (c) the ad hoc task-specificity of ANN architectures, and (d) the high energy demands of running the algorithms on computers. Similarly, despite their biological inspiration at an abstract level, ANNs lack many of the computational operations that the known neuronal biophysics implies. It is conceivable that this latter disparity in implementation also underlies the former mismatch in performance. In particular, much of the complexity within biological neurons (a) is dedicated to synaptic plasticity for learning, (b) is governed by dynamically changing chemical concentrations, (c) is maintained across brain areas and species, and (d) accomplishes extreme energy efficiency. It is hard to ignore that these four biological properties that are largely missing from ANNs have a one-to-one correspondence to the four aforementioned limitations of ANNs. Based on this, our high-level ambition here is to explore whether these biological properties and the associated advantages can be brought to Deep Learning by a new rendition of a particular neuromorphic mechanism, i.e.\ short-term plasticity (STP) of synapses.
\begin{figure}
	\centering
	\includegraphics[width = .5\textwidth]{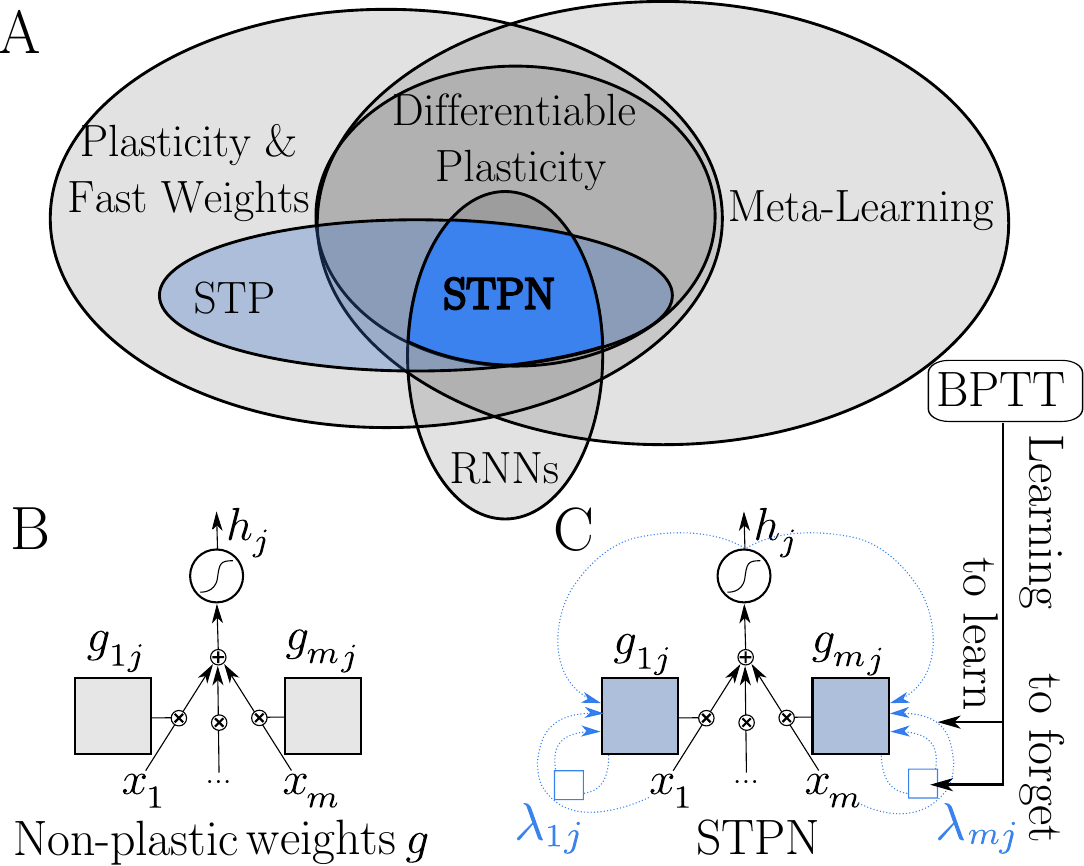}
	\caption{(A) Our STPN model in the literature's context. (B) Simple connectionist neuron, with synaptic efficacies g. (C) The STPN. Plasticity through recurrent parameters, e.g.\ $\lambda$.}
	\label{fig:venn}
\end{figure}
Our approach is closely related to concepts from multiple subfields (\cref{fig:venn}.A), which we review in this section.

\subsection{Neuromorphic Computing}
Our goal here is aligned with the field of neuromorphic computing, which has been porting biophysical mechanisms from experimental neuroscience into models and emulating them in electronic circuits to improve neural computation \cite{indiveri2011frontiers,indiveri2021introducing, sarwat2022phase, sarwat2022chalcogenide}. However, the focus has been mostly on the specific property of spiking neuronal activations, in so-called spiking neural networks (SNNs) \cite{maass1997networks, ponulak2011introduction}. In addition, the field's aim has been overwhelmingly to increase energy \textit{efficiency} of the state of the art (SOTA), rather than its \textit{proficiency} such as classification accuracy or achieved reward in useful tasks. Nevertheless, recent results show that advantages of neuromorphic mechanisms are not limited to efficiency. If other properties than spikes are considered, such as STP, neuromorphic models can in fact also lead to more proficient models \cite{moraitis2020optimality}, while remaining compatible with efficient electronics \cite{sarwat2022phase}.

\subsection{Plasticity and Short-Term Plasticity}
The term ``plasticity" refers to the rules that determine how the efficacy, i.e. strength, of synaptic connections changes in the brain or in model networks. The term is often reserved for local rules, i.e. when the changes of a synapse depend on signals from the pre-synaptic and/or the post-synaptic neuron that the synapse connects, and potentially a third signal, such as concentration of neuromodulators \cite{gerstner2018eligibility, pogodin2020, sarwat2022chalcogenide}. Such plasticity is generally considered the underlying principle of learning in the brain, and a possible path to life-long ML. One type of plasticity rule is STP \cite{zucker1989short, tsodyks1997neural, chamberlain2008role, mongillo2008synaptic}, i.e.\ a type of plasticity with strong biological evidence, whose effect is constrained in time. On the other hand, if each plastic change is persistent, the rule is a type of long-term plasticity. For example, Hebb postulated that biological weights are updated proportionally to pre- and post-synaptic activations \cite{hebb1949organisation}. Despite its simplicity, networks with Hebbian-like plasticity and without supervision can be used to optimize models such as Bayesian mixtures \cite{nessler2009stdp} and to solve tasks such as handwritten digit classification fast and with robustness to adversarial attacks \cite{moraitis2021softhebb}.
Extensions of such plasticity can be both insightful and powerful, playing a key role in the ML SOTA \cite{nessler2009stdp, scellier2019equivalence, lowe2019putting, limbacher2020hmem, millidge2020predictive, illing2021local}.
Particularly relevant to the present manuscript are cases where a plasticity rule causes synaptic changes at a different timescale with respect to another learning rule. This concept of ``subordinate" or ``interleaved" changes in synapses has been termed dynamic weights \cite{feldman1982dynamic}, fast weights \cite{hinton1987using, schmidhuber1992learning, schmidhuber1993reducing, tieleman2009using, ba2016using, schlag2017gated}, or simply plasticity or learning \cite{bengio1990learning, moraitis2018spiking, miconi2018differentiable, miconi2020backpropamine, moraitis2020optimality, miconi2021learning}. Interestingly, some of these approaches that use associative plastic updates have been shown to be equivalent to attention mechanisms \cite{ba2016using}, and even to models like linear transformers \cite{schlag2021linear}.

The short-term aspect of STP can be modelled by splitting the synaptic efficacy $G$ that weights a presynaptic input, into a long-term weight $W$ and an additive or multiplicative short-term component $F$, e.g.\ $G=W+F$. Subsequently, an update rule, which depends on local variables, increments $F$, which otherwise decays exponentially with time, implying a type of learning followed by forgetting. 
The literature has shown diverse functions emerging from the various forms of STP. For example, STP can apply temporal filtering on synaptic inputs \cite{rosenbaum2012short}, or underpin biophysical models of working memory \cite{mongillo2008synaptic, szatmary2010spike, fiebig2017spiking}. If realized as the fast-weight counterpart to a concurrent slower plasticity, it can focus learning from sequences on multiple timescales of the input \cite{moraitis2018role, moraitis2018spiking}. It also instils a long short-term memory to recurrently connected neural networks (RNNs), resulting in properties and performance similar to Long Short-Term Memory (LSTM) \cite{hochreiter1997long} units \cite{Bellec2018}. In all these cases, STP was studied in the context of SNNs. The reason is that the increased biological plausibility of the spiking activations is a more useful model for many neuroscientists, and may bring energy efficiency advantages to neuromorphic engineering. However, SNNs are also harder to analyse mathematically or train practically, which limits the potential improvements from STP in useful ML tasks.

Nevertheless, two recent models with STP did outperform others without in specific tasks, even though the STP-equipped models were SNNs. Namely, first, STP led to improved learning of restricted Boltzmann machines from unbalanced data \cite{leng2018spiking}. 
Second, and in closer relation to our present study, a very simple SNN learned without supervision from the standard static MNIST dataset of handwritten digits \cite{lecun1998gradient}, but was tested on the classification of frames of a video of digits with moving occlusions \cite{moraitis2020optimality}. Surprisingly, owing to STP's dynamics at the input synapses, the SNN outperformed simple supervised convolutional neural networks and LSTMs, even trained on the video dataset with temporal context.
\citet{moraitis2020optimality} also included a mathematical proof that neural networks with STP at their input synapses are in fact the optimal model for certain dynamic data. The key enabling principle is that input synapses with STP memorize not only dataset-wide features in the long-term weights $W$, but also recent features that are relevant to the immediate future in the short-term component $F$. However, the practical demonstrations of STP's advantages remain limited due to the reliance on spiking neurons.
It should be noted that some of the non-spiking models with fast weights, particularly those co-authored by Hinton \cite{hinton1987using, tieleman2009using, ba2016using}. \citet{hinton1987using, tieleman2009using} did include STP-like decaying dynamics in the fast subordinate weight changes, however training of the STP's parameters was not reported. In \citet{ba2016using}, backpropagation trained an RNN's conventional slow weights end-to-end, while additional fast weights were updated by a separate non-learnable rule, also during inference. This system was able to solve tasks where attention to the recent inputs is important. However, the authors did not include STP in input synapses, but only in recurrent connections between hidden units.
Here, we hypothesize that STP's potential for Deep Learning can be realized by enabling STP also at the input synapses, as in the theoretically supported proposal of \citet{moraitis2020optimality}.

In order to fulfil the promise of optimally adapting plasticity, one important element is missing from those previous studies. That is the optimization of the STP rule for each synapse and through learning, as opposed to fixed STP based on chosen hyper-parameters and uniform for all connections. How to realize this last ingredient is not immediately obvious.
One recent work \cite{tyulmankov2022meta} did train a model equipped with the Hebbian STP mechanism of \citet{moraitis2020optimality}. However, that model does not support recurrent connections between neurons, uses uniform plasticity parameters across all synapses, and focuses on simple associative tasks of random binary inputs. Therefore, the key advance was that the uniform hyperparameter in the earlier work was hand-tuned as opposed to optimized through backpropagation.
Here we target advanced tasks, through fully recurrent models, and through the training of individual synapses' STP. To achieve this, we implement STP in a synapse as a sub-connection within that synapse, thus resulting in a formulation of STP Neurons (STPNs) as a novel type of recurrent unit (see \cref{sec:stpn}). Through this formulation, the learning-and-forgetting function of STP becomes itself trainable, drawing links to the category of algorithms that are meta-learning, i.e.\ learning to learn.


\subsection{Learning to learn}
Meta-learning is a paradigm that applies machine learning to improve further learning in new domains. This has been considered analogous to the nesting of biological timescales of evolution, development, life-long skill learning, and learning for temporary objectives. In fact, direct empirical evidence for meta-plasticity, i.e.\ changes in plasticity, in the brain has been observed \cite{abraham1996metaplasticity}. More generally, a large body of ML literature on meta-learning exists, with various approaches and applications \cite{schmidhuber1996simple, thrun1998learning, Bellec2018, hospedales2020meta}. For instance, it has been shown that, through minimal modifications to how the training data is provided, RNNs learn to learn, where the inner learning loop consists in changes of recurrent state rather than changes of weights \cite{Hochreiter01learningto, wang2016learning}. Clearly, some of the most relevant meta-learning methods with respect to our work here are those where learning to learn consists in learning the parameters of a plasticity rule that controls the changes of weights within the inner loop. Both evolutionary \cite{soltoggio2018born} and gradient-based \cite{bengio1990learning} algorithms have long been described for such plasticity-rule meta-learning purpose. However, backpropagation-based end-to-end training of the individual-synapse plasticity parameters along with the other parameters of a neural network has only recently been demonstrated \cite{miconi2018differentiable}. The trainable plasticity in that work outperformed non-plastic neural networks and has been followed up with extensions that confirm its advantages \cite{miconi2018backpropamine, beaulieu2020learning, miconi2021learning}. However, thus far, none of these has incorporated STP. It has not been obvious how to learn the spontaneous temporal dynamics of synaptic efficacy decay, i.e.\ how to learn the forgetting aspect of STP, for each synapse. As a matter of fact, in models without STP, fast weights are long-term, i.e.\ they persist through time, unless the fast-weight-update mechanism applies a learning increment or decrement. As a result, they lack a dedicated forgetting mechanism, and forgetting must be handled by the mechanism intended for learning.
We believe that demonstrating the importance of learning to forget would be an important contribution to the meta-learning field, as it could possibly relate to the challenge of catastrophic forgetting in continual deep learning. Catastrophic forgetting refers to the phenomenon of neglecting a previously learned task because of learning a new one, and it is one of the important motivations for meta-learning research \cite{vuorio2018meta, ren2018incremental, flennerhag2019meta, javed2019meta, hospedales2020meta, Sinitsin2020Editable}. It is plausible that learning, not only to learn, but also to explicitly forget would mitigate catastrophic forgetting. Here indeed we present a procedure of learning to learn and forget, realized as learning of STP parameters.

%
%

\subsection{Contribution to the field}
We present STPN, a new recurrent type of unit that expands the family of RNNs with the possibility of a recurrent state within each input synapse. It extends other fast-weight models by adding STP to inputs, and by making STP trainable per synapse. It builds on other models of differentiable plasticity by including a short-term aspect. It complements learning to learn with learning to forget. We will show that it is a better RNN choice than LSTM, surpasses the most recent fast-weight models, and outperforms other differentiable plasticity mechanisms, in a variety of tasks, with supervised and reinforcement learning, including examples of meta-learning. STPN's benefits comprise both improved task proficiency and energy efficiency.

\begin{figure*}[h]
	\centering
	\includegraphics[width = \textwidth]{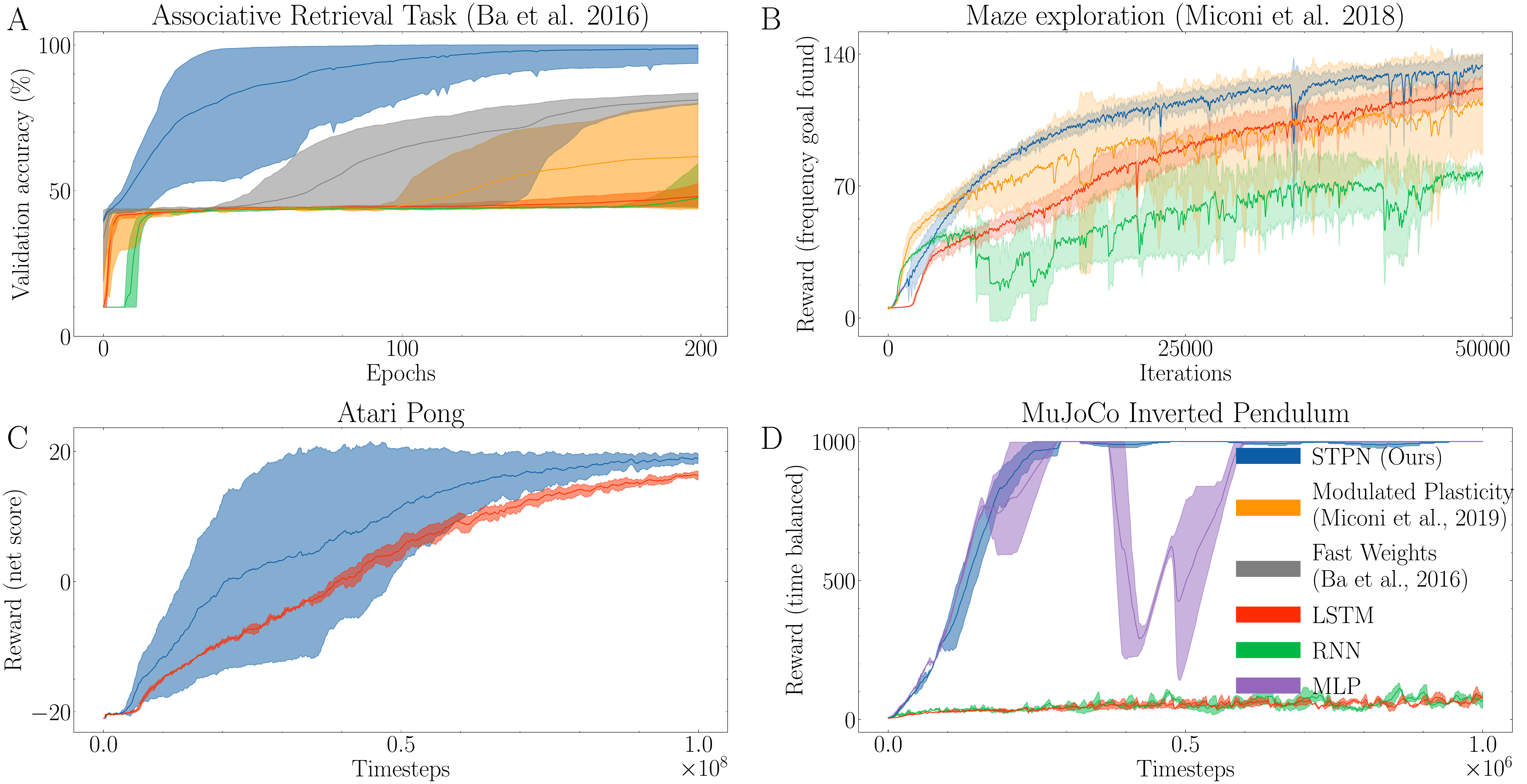}
	\caption{Proficiency during training of the STPN and other models across 4 tasks. Proficiency refers to: (A) Validation accuracy in ART \cite{ba2016using};  (B) Accumulated reward in an episode of the Maze Exploration task \cite{miconi2018differentiable}; (C) Net score in the Atari game Pong \cite{bellemare2013arcade} when one of the players reaches 21. (D) Number of timesteps in balance within an episode, capped at 1000, in Mujoco Inverted Pendulum \cite{todorov2012mujoco}.}
	\label{fig:proficiency}
\end{figure*}

\section{The STPN model}
\label{sec:stpn}
\subsection{The model}
We begin to construct the STPN model by defining initially a simple feed-forward hidden unit with activation $h_j$, that receives an input vector $\boldsymbol{x}$ from a preceding layer, and weights it by a synaptic efficacy vector $\boldsymbol{g}_j$ (\cref{fig:venn}B). Two realizations (\cref{fig:venn}C) serve as intuition for casting such a model as a recurrent unit when STP is added. Firstly, the postsynaptic activation $h_j$ may feed back to the input synapses as one of the factors of the plasticity rule, e.g if STP is Hebbian. Secondly, the short-term memory and decay of each efficacy $g_{ij}$ implies that the synapse contains a state variable that is partly propagated to itself forward in time. Intuitively, these are two self-recurrent loops. 

The specific type of STP that we choose to base the STPN model on is the one that was motivated theoretically in \citet{moraitis2020optimality}. According to this, a hidden neuron $j$ receiving an input vector $\boldsymbol{x}$ from a preceding layer, weights it by a synaptic efficacy vector $\boldsymbol{g}_j$ that consists of two additive components: $\boldsymbol{g}_j=\boldsymbol{w}_j+\boldsymbol{f}_j$.
STP acts on $\boldsymbol{f}_j$. Meanwhile, component $\boldsymbol{w}$ may be fixed, or updated by a different learning rule, either concurrently with STP, or preceding it. In this work, $\boldsymbol{w}$ is updated via backpropagation in the outer loop between sequences. This STP rule dictates that $f_{ij}$ in a synapse $j$ from neuron $i$, firstly decays exponentially with time $t$, and secondly is subject to Hebbian plasticity proportional to the pre- and post-synaptic variables $x_i$ and $h_j$.

Based on the intuition for the two self-recurrent loops we construct STPN as a recurrent unit that aims to have analogous functionality to the latter Hebbian STP rule. We introduce upper-case symbols $\boldsymbol{G}=\boldsymbol{W}+\boldsymbol{F}$ to indicate the matrices that connect the input vector to the hidden unit vector, and two equally-sized matrices $\boldsymbol{\Lambda}$ and $\boldsymbol{\Gamma}$ that contain elements $\lambda_{ij}$ and $\gamma_{ij}$, which parameterize the STP. We will symbolize the element-wise and outer products by $\odot$ and $\otimes$ respectively.

One time-step's pass through a layer of STPN units in this basic version, and assuming fixed weights $\boldsymbol{W}$ and a non-linearity $\sigma(\cdot)$, is described by the following set of equations, where $\mathds{1}$ indicates a matrix of ones:
\begin{gather}
	\boldsymbol{G}^{(t)}=\boldsymbol{W}+\boldsymbol{F}^{(t)} \label{eq:split_weights}\\
	\boldsymbol{h}^{(t)}=\sigma(\boldsymbol{G}^{(t)} \boldsymbol{x}^{(t)})
	\label{eq:activation}\\
	\boldsymbol{F}^{(t+1)}=\boldsymbol{\Gamma} \odot (\boldsymbol{x}^{(t)} \otimes \boldsymbol{h}^{(t)})+(\mathds{1}-\boldsymbol{\Lambda}) \odot \boldsymbol{F}^{(t)}\label{eq:increment}
\end{gather}
The activation $\boldsymbol{h}$ depends on the short-term component $\boldsymbol{F}$ through $\boldsymbol{G}$ (\cref{eq:split_weights,eq:activation}). $\boldsymbol{F}$, in turn, depends on the activation and on itself (\cref{eq:increment}), wherein lies the recurrency of this model. Unlike in standard RNNs, this recurrency does not necessitate recurrent connections between hidden states $h_j$, i.e.\ there are no synapses connecting units of the same layer. The recurrency is mediated in this case through the parameter matrices $\boldsymbol{\Gamma}$ and $\boldsymbol{\Lambda}$ that connect the synaptic state to itself and to its postsynaptic neuron. Therefore, this model realizes an uncommon type of recurrent connectivity, characterized by meta-weights, i.e.\ connections (self-loop from $f_{ij}$ through $\lambda_{ij}$ in \cref{fig:venn}D) within each synaptic connection (blue square $g_{ij}$ in \cref{fig:venn}D).

In meta-learning terms, the Hebbian STP of Eqs. (\ref{eq:split_weights})-(\ref{eq:increment}) describes one iteration of an inner learning (and forgetting) loop that is unsupervised (see \cref{sec:L2L_wSTPN}).


\subsection{Equivalence to STP}
It can be shown rather simply that the STPN construct is not merely analogous to a neuron with the Hebbian STP rule that we focus on, but it is exactly this - however in the discrete rather than the continuous time domain. The original equations of the rule dictate firstly an exponential decay over time with a rate $0<\lambda_{ij}<1$:
\begin{equation}
	\label{eq:decay_cont}
	\frac{df_{ij}}{dt}^{(decay)}=-\lambda_{ij} f_{ij}.
\end{equation}
Secondly, at any discrete time point that the synapse receives an input $X_i^{(t)}$, Hebbian plasticity with a learning rate $\gamma_{ij}$ increments $f_{ij}$ as well, by a $\Delta f_{ij}$ that also depends on the post-synaptic output $h_j^{(t)}$:
\begin{equation}
	\label{eq:hebb_cont}
	\frac{df_{ij}}{dt}^{(Hebb)}= \Delta f_{ij}^{(t)(Hebb)} = \gamma_{ij} x_i^{(t)} h_j^{(t)}.
\end{equation}
The combined effect of \cref{eq:decay_cont,eq:hebb_cont} describes the full original STP rule.
The continuous-time \cref{eq:decay_cont} can be approximated arbitrarily closely by small discrete time-steps through the Euler method \cite{euler1794, kendall1989introduction}:
\begin{equation}
	\label{eq:decay_discrete}
	\delta \boldsymbol{F}^{(t+1)(decay)}=-\boldsymbol{\Lambda} \odot \boldsymbol{F}^{(t)}.
\end{equation}
This method has long been used to simulate the continuous-time evolution of neuromorphic models, such as spiking neurons, including recently \cite{wozniak2020deep}.

On the other hand, \cref{eq:hebb_cont} is already only dependent on discrete-time events, so in discrete time it remains the equivalent:
$	\label{eq:hebb_discrete}
	\Delta\boldsymbol{F}^{(t+1)(Hebb)} = \boldsymbol{\Gamma} \odot (\boldsymbol{x}^{(t)} \otimes \boldsymbol{h}^{(t)}).
$
STPN's \cref{eq:increment} can therefore be written as 
\begin{equation}
	\boldsymbol{F}^{(t+1)}=\boldsymbol{F}^{(t)}+\Delta\boldsymbol{F}^{(t+1)(Hebb)}+\delta \boldsymbol{F}^{(t+1)(decay)},
\end{equation} 
which shows that the model indeed is equivalent to discrete-time Hebbian STP.

\subsection{Learning to learn and forget with STPN}
\label{sec:L2L_wSTPN}
STPN's parameters $\gamma_{ij}$ and $\lambda_{ij}$ are interpretable quite concretely in two ways. Firstly, as parameters of recurrent connections, they have a clear role of short-term memory. Secondly, $\gamma$ is also the Hebbian learning rate of the synapse, whereas $\lambda$ is its forget rate. Further, by framing the model as a network that is based on standard recurrent weighting operations, STPN's parameters are trainable. That is, not only the standard between-neuron connections $\boldsymbol{W}$, but also the characteristic connections-within-synapses can be trained. Notably, as these latter learned connections act as STP's rates of learning and forgetting, then training these parameters realizes a learning-to-learn scheme that also learns to forget. In this realization of meta-learning, the inner learning loop is the online unsupervised adaptation of the network through STP (\cref{eq:increment}) to a given input sequence, whereas the outer learning loop consists in the optimization of the network's parameters, e.g.\ via backpropagation through time (BPTT), for the inner online learning task, over multiple examples of this inner task (see \cref{alg:example} in the Appendix).

\subsection{STPN variants}
Even though thus far we have provided a description of STPN based on a feed-forward base structure (\cref{eq:activation}), the same connection-within-a-synapse type of recurrency and plasticity can also be added to models that have recurrency between hidden units, like simple RNNs do. This simply changes the unit to also receive inputs from the same hidden layer, i.e.\ changes \cref{eq:activation} into 
\begin{equation}
	\boldsymbol{h}^{(t)}=\sigma(\boldsymbol{G}^{(t)} [\boldsymbol{x}^{(t)}; \boldsymbol{h}^{(t-1)}]), \label{eq:activation_r}
\end{equation}
accompanied by the corresponding change in size of the parameter matrices $\boldsymbol{F}$ and $\boldsymbol{W}$. We refer to the networks of \cref{eq:activation,eq:activation_r} as the STPNf and STPNr variants, respectively, of STPN connectivity (for their \textbf{F}eed-forward and \textbf{F}ully connected, or \textbf{R}NN non-plastic skeleton). This formulation is extensible to further variants, such as STPNl for an STPN with the addition of LSTM's gating, although those are not explored in this work. The results in \cref{sec:exps} are achieved using the STPNr in most tasks, except for a fully-observable robotics environment where we tested the STPNf.

\begin{figure*}[!h]
	\centering
	\includegraphics[width = \textwidth]{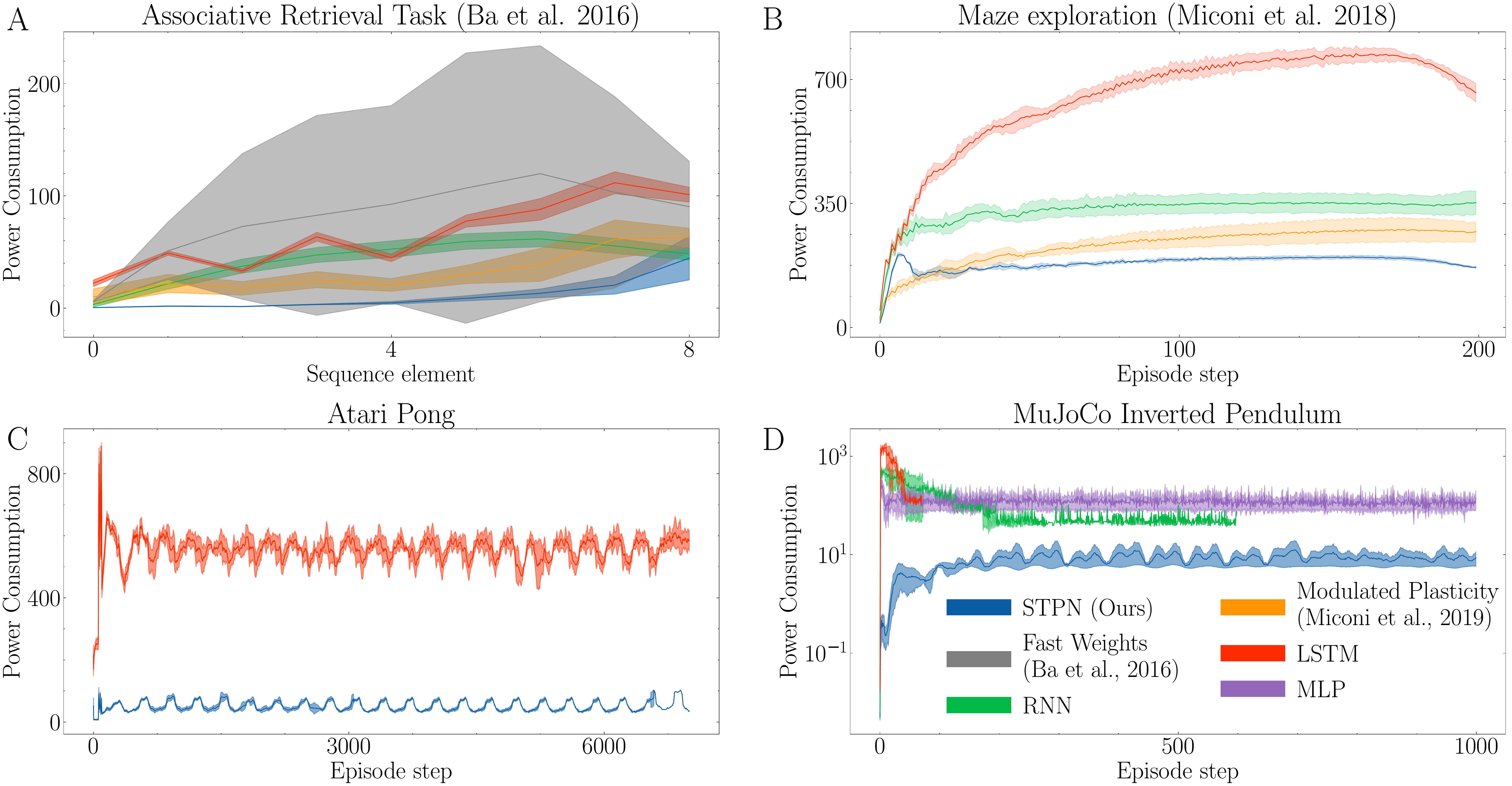}
	\caption{Power consumption of trained models during inference, evaluated at each timestep. Mean and standard deviation over multiple random seeds are shown, averaged over multiple inference runs. Lower is better.}
	\label{fig:efficiency}
\end{figure*}
\section{Methods}
\label{sec:methods}

\subsection{Training methods}
We found that, given the complexity of our model's per-synapse STP parameters, and especially for the more complex STPN models that include recurrency both in the synapse and between neurons such as the STPNr, training stability and robustness are not possible automatically. We explored approaches for making the STPN's training robust, and arrived in a strategy that consists in an initialization method and a weight-normalization approach that we introduced. See Appendix \cref{alg:example} and \cref{sec:app_trainingmethods} for details.

\subsection{Energy consumption measurement}
Our approach focuses on the synaptic weighting mechanism, and as such it could provide guidelines for future neuromorphic hardware. Its energy consumption is therefore a key concern. Given that STP is a highly neuromorphic mechanism inspired by biophysics, it is probable that it is also highly efficient. In principle, STP has the potential to depress synaptic currents flexibly and independently through Hebbian decrements and short-term decay, which is a reason to intuitively expect high energy-efficiency from STPN. With this in mind, we measured the power consumption that each model would incur through weighting operations in hypothetical analog neuromorphic hardware. In such hardware, presynaptic input $\boldsymbol{x}$ is provided as a voltage $V$, whereas the matrix $\boldsymbol{G}$ of synaptic efficacies $g$ could be represented in an array of resistive devices with conductance $g$, such as memristors \cite{sarwat2022phase}. This enables weighting and summation through Ohm's and Kirchhoff's laws, and incurs a power consumption by each synapse, which is proportional to the conductance $g$ and the square of the applied input voltage $V^2$, as shown in \cref{eq:power-consumption-derivation}
\begin{align}
	\label{eq:power-consumption-derivation}
	P = V I = V \left(\frac{V}{R}\right) = V^2 \left(\frac{1}{R}\right) = V^2 g,
\end{align}
where I is the electrical current, and R is the resistance.
This method allowed us to measure the hypothetical power consumption of the various models using $P = x^2|g|$.
Notably, analog signals underlie synaptic transmission also in biological synapses, through ions. Fewer ions and smaller currents are transmitted when the biological synapse is depressed. Therefore, to the extent that STP is biologically plausible, our measurements of the STPN's power consumption provide some insight also into STP's role in the brain's energy budget.

\subsection{Tested baseline models}





In our experiments, we compare the STPN with other widely used networks with memory, like standard RNN and LSTM, and others with different kinds of synaptic memories, like RNN with Fast Weights \cite{ba2016using} and Modulated plasticity RNN \cite{miconi2018backpropamine}. Given the qualitative architectural differences across the tested models, we make them comparable by choosing hidden sizes that correspond to equal number of parameters between models in each experiment. 
\citet{ba2016using} augmented RNNs by adding \textbf{Fast Weights} to some connections, specifically between recurrent neurons, i.e.\ not in feedforward synapses such as from the input layer. Also, the hyperparameters of plasticity are uniform throughout all fast-weight synapses: Hebbian update factor and decay of current memories. Additionally, they describe the use of an inner loop where fast weights can repeatedly act on an intermediate hidden state and be updated, while the effect of slow weights acts as a sustained boundary condition in each iteration. However, they found no significant empirical benefit in performing multiple such iterations; and neither do we when tuning this baseline. Additionally, note that this same mechanism can also be applied to any other network with synaptic memories, and is conceptually closer to work on recurrent processing of the same input as a form of adding computational resources \cite{schwarzschild2021can,schwarzschild2021uncanny,banino2021pondernet}. Fast Weights RNN is a conceptually simpler version of STPN, primarily due to its use of plasticity only on recurrent connections, the plasticity rule being uniform across neurons and synapses, and the fact that parameters modulating fast learning and forgetting are not trained (but tuned using the validation set). Other mechanisms implemented by the Fast Weights network, like layer-normalization of the hidden activations, and a repeated application of the plasticity through multiple recurrent iterations on each time-step of the input sequence. Our control experiments (not shown) showed that this iterative aspect offers only a small improvement, and could be equally applied to STPN. Therefore, the Fast Weights model serves as a good baseline to the training of a synapse-specific STP mechanism. 
\citet{miconi2018differentiable} exploit the idea of optimizing plasticity by training parameters that control each synapse's memory via backpropagation, hence \textbf{Differentiable Plasticity}. \citet{miconi2018backpropamine} further add a modulating term to the plasticity, akin to a third factor of three-factor plasticity rules, which is also trainable. Unlike STPN, in that prior work no variant could evolve (e.g.\ decay) its efficacies spontaneously with time, like STPN does through $\lambda$. We select the modulated plasticity variant (``Modplast'') as a baseline due to its superior performance among the variants tested in the Maze Exploration task in \cite{miconi2018backpropamine}. The other two major differences in this specific model with respect to STPN are: a) plasticity only in recurrent synapses, b) a meta-trained parameter per synapse that clips the plastic part of the efficacy. This model being the best out of multiple variants of trainable plastic RNNs serves as a good baseline for the STPN.
\textbf{HebbFF} \cite{tyulmankov2022meta} is architecturally equivalent to STPNf with uniform plasticity. However, it offers no option for per-synapse plasticity parameters, recurrent between-neuron connections, or mechanisms to stabilize plastic updates such as the ones that we introduced \cref{sec:methods}. These become necessary in tasks with higher complexity than those tested in \cite{tyulmankov2022meta}, which we tested and confirmed. Specifically, given that HebbFF could be described superficially, as a highly-simplified version of the STPN, we include comparisons to HebbFF alongside several other simpler STPN versions in \cref{ssec:Appendix:per-synapse-ablation}.
We also carry out experiments using networks with non-synaptic memories, but which are general purpose ANNs with memory, and shown to perform meta-learning through the encoding of their neuronal memories \cite{wang2016learning}. We compare the performance of \textbf{RNN} (a direct comparison to STPNr but without synaptic STP, and a different type of memory to STPNf) and \textbf{LSTM} (as a more complex non-plastic RNN with gating and an additional memory mechanism).

\begin{figure*}[h!]
	\centering
	\includegraphics[width = \textwidth]{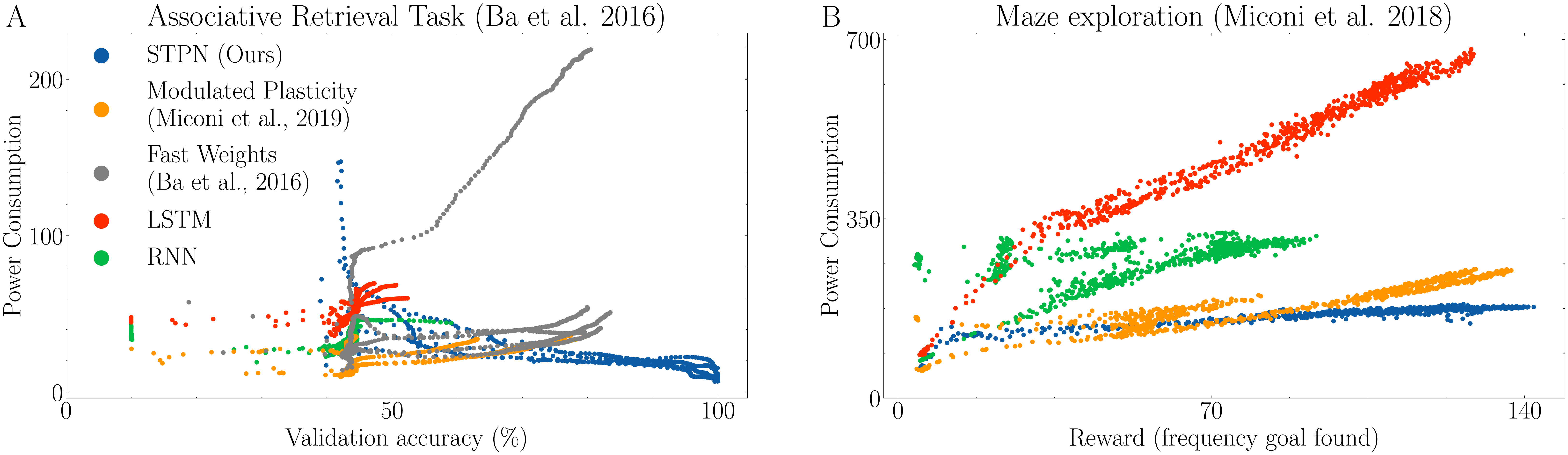}
	\caption{Energy efficiency as a function of task-specific proficiency, measured at regular intervals during training. Each scattered point is one instance of the model at a specific point of learning. All seeds for a model shown with the same color.}
	\label{fig:efficiency_vs_proficiency}
\end{figure*}

\subsection{Tested tasks} \label{ssec:methods:tasks}




\textbf{Associative Retrieval Task} (ART) \cite{ba2016using} tests the capabilities of a network to successfully store associations between pairs of elements seen in a sequence, and retrieve one of the elements of the pair (the value) when queried with the other (the key) at the end of the sequence. This task and other variants \cite{schlag2017gated, le2020self} are therefore commonly used to compare the abilities of networks with memories of different nature. For the setup of this task, we mostly follow the experimental details provided in \citet{ba2016using}, besides the modifications described in \cref{sssec:details-art}. 
\textbf{Maze Exporation:} Maze or grid-like tasks have been commonly used in RL as a data efficient and interpretable task to test RL algorithms. \cite{miconi2018differentiable} instantiates a specific form of such maze, in which an agent has an egocentric view of its surroundings and whose goal is to navigate towards a reward, which it cannot see and whose position is randomized across episodes. Furthermore, the agent is randomly relocated to another position in the grid upon reward-finding and at the start of an episode. Given this environment design, we can think of the Maze Exploration task as a test to the meta-learning capabilities of the agents. Firstly, each maze instantiation (with a fixed reward position) represents the inner loop of learning, as the agent needs to learn how to effectively navigate towards this new reward position, and do so from each initial random position after relocation when the reward is found. Secondly, across different episodes, parameters need to be optimized as to help within the inner loop and through different instantiations of the maze, hence they are learned to learn. Additionally, even within an episode, we can distinguish two qualitatively different phases: in the first steps, the agent needs to explore this instantiation of the maze in order to find the reward location. Upon learning of the positioning of the goal, it changes to an exploitation phase in which the key is to quickly reach the already seen reward. These two task distributions (the distribution of all possible mazes and the binary exploration-exploitation phases) positions this task as a good benchmark for the meta-learning and sequence adaptiveness capabilities of non-plastic and plastic networks with memory.
\textbf{Atari games and MuJoco simulated robotics:}
The previous two experimental setups, namely ART and Maze Exploration, were presented in plasticity related articles and have some characteristics that might favor models with plastic connections. RL is an area of research where the use of RNNs and its variants is still widespread in SOTA algorithms, more so in comparison to supervised learning domains regarding language, video or audio. In order to go beyond simple tasks to which the plastic networks literature has been limited, in the interest of exploring a domain in which RNNs are a component of SOTA approaches, and with the additional intent to examine the use of STPN as a general-purpose network with memory, we present some preliminary experiments with two tasks from two common Deep RL benchmarks: Atari Pong and MuJoCo Inverted Pendulum. 
\textbf{Pong }is an Atari 2600 game implemented in the Arcade Learning Environment (ALE) \cite{bellemare2013arcade} in which the player faces an opponent, each player controls a bar that they can move up and down in order to hit a ball and therefore score a goal or stop the opponent from doing so. The game stops when a player reaches a score of 21. The reward obtained by the agent is the net score of the finished game, and the observations are the game frames. For our experiments in Pong, we use A2C, the synchronous version of A3C \cite{mnih2016asynchronous} as the learning algorithm. The agent's network uses convolutional layers such as those in  \citet{mnih2016asynchronous}, a recurrent policy (shared by value and action branches) and non-plastic, feed-forward value and action branches. A simple algorithm such as A2C allows to better explore the real impact of the recurrent unit, and its previous use in the Maze task provides us with some assurances of its compatibility with plastic networks. Beyond the shift of domain from toy, plastic-network favoring tasks, Pong offers the first situation in which STPN is embedded in a larger network, and whose input is not sparse and directly from the environment, but dense and the result of the processing of multiple layers. Note because we do not perform frame stacking, the environment is a partial observable Markov decision process (POMDP), given that velocity of the ball cannot be inferred from a single frame \cite{hausknecht2015deep}. This need for agents with memory justifies the use of recurrent policies and lack of comparison with a feed forward policy. MuJoCo \cite{todorov2012mujoco} is a physics engine widely used for research in robotics and reinforcement learning.
\textbf{Inverted Pendulum} is one of the simplest tasks within MuJoCo, where the agent aims to balance an inverted pendulum which sits on a cart, by moving the cart forward and backward in one dimension with continuous valued controls. Our experiment with MuJoCo's inverted pendulum represents a preliminary step in the direction of using STPN in more complex problems in robotics and continuous control like those in the MuJoCo environment, and in combination more advanced RL algorithms, like Proximal Policy Optimization (PPO) algorithm \cite{schulman2017proximal} in this case. Unlike Pong without frame stacking, InvertedPendulum is a fully observable Markov decision process (MDP), so memory of the past is not needed to assess the present or predict the future. Therefore, training the parameters that process hidden memories could represent an unnecessary effort in terms of credit assignment, aggravated by hidden states being of much greater dimensional that environmental observations. Therefore, we use the STPNf in this case, and MLP as a baseline; in addition to RNN and LSTM.
For the \textbf{terminology} we use for different timescales (e.g.\ episodes, epochs, timesteps), see \cref{sec:app_term}.


\section{Results}
\label{sec:exps}

\subsection{Accuracy \& reward}
%

\cref{fig:proficiency} shows that the STPN is more proficient, i.e.\ obtains larger validation accuracy and reward, than all other baselines. These plots report the mean and standard deviation over multiple seeds, except in \cref{fig:proficiency}.C , where we follow recommendations given in \citet{agarwal2021precipice} and report within 0.25 inter-quantile ranges, due the known propensity to certain catastrophic and non-representative runs of brittle algorithms like A2C in ALE. Furthermore, we find the unsupervised adaptability of STPN's plastic weights offers greater performance at inference time, as detailed in \cref{table:proficiency}. \cref{ssec:Appendix:per-synapse-ablation} presents the empirical advantages of including recurrent connections and learning per-synapse plasticity for STPN, with respect to non-recurrent or uniform plasticity versions of STPN. Additionally, we present benefits over HebbFF \cite{tyulmankov2022meta}, a simpler feed-forward STPN with uniform plasticity.

Across all tasks, we observe STPN show noticeable stability in the learning process while achieving higher proficiency earlier in training. This is particularly clear in the Inverted Pendulum task, where the mid-training instability of MLP, also reported by \citet{schulman2017proximal}, is counteracted by the addition of STP to the synapses. We hypothesize that compared to other plastic models, explicitly learning decaying dynamics of synaptic memories helps adding stability to processing of the same inputs across time, as less relevant synapses will be closer to their long-term value $w_{ij}$.

STPN seems to be more robust to hyperparameter choice than other tested baselines. Three examples are: a) the fact that it outperforms modulated plasticity, a similarly complex plastic model, with the same parameters that are tuned for that specific network; b) in Pong, STPN can tolerate a larger learning rate than LSTM, as the latter was unable to learn with high learning rate for most random seeds, such that it needed a lower learning rate for its best mean final reward; and c) like in Maze Exploration, STPN outperforms the model for which the experiment hyperparameters are tuned in Inverted Pendulum.

%

\subsection{Energy consumption}

The initial hypothesis regarding the trainable dynamical suppression of synapses, specially to input synapses, improving the energy efficiency of backpropagated-trained models is unequivocally confirmed in our experiments, as shown in \cref{fig:efficiency,fig:efficiency_vs_proficiency}. See also Appendix \cref{table:efficiency}. Unlike for proficiency metrics like accuracy or reward, the networks are not given any explicit signal or instruction to be more energetically efficient, making these results even more remarkable. However, in some cases a common goal can improve both proficiency and efficiency in the STPN: suppress unimportant synapses. We believe these results show such a goal is met. This is particularly clear in \cref{fig:efficiency_vs_proficiency}. (A) As the network optimizes proficiency, its energy consumption decays. Similarly, in \cref{fig:efficiency_vs_proficiency}. (B) The growth of energy with respect to reward is minimal relative to other models, and largely attributable to past reward being part of the agent's observation in this experiment (hence a higher reward leading to a higher norm in the input and higher energy consumption).

\subsection{Learning to learn and forget}
Standard RNNs are seen as being able to meta-learn the encoding of a learning algorithm within its neuronal memories \cite{wang2016learning}. STPN adds a synaptic state, which is updated by three meta-learned parameters (the long term synaptic efficacy, and the two STP parameters $\lambda, \gamma$), and has the potential of a higher influence in the output of the unit as compared to neuronal memories. The higher proficiency achieved by STPN on meta-learning tasks shows a learning-to-learn process is improving the predictive performance of the network. Further evidence for learning to learn and forget in the STPN is the following. Firstly, high proficiency is reached early in training, without incurring in too many instabilities or preventing further optimization at later stages of training, as can be seen in \cref{fig:proficiency}. Secondly, the negative slope of the efficiency vs proficiency curve of the STPN scatter points (\cref{fig:efficiency_vs_proficiency}) indicates a learning-to-forget mechanism, that improves both aspects.

\section{Discussion}
STPN is founded on prior theoretical optimality proofs and biological evidence. As a result, STPN is more performant, efficient and learns to better learn and forget than a variety of other models, and in a broad range of difficult tasks, by introducing individually trainable short-term plasticity to all synapses. The model also offers a new method to increase the efficiency of neuromorphic platforms. In combination with the prior experimental evidence that supports the biological plausibility of our plasticity, our results also raise STP's significance for animal behaviour and for the brain's energy budget. This outcome paves new and interdisciplinary research avenues, in RNNs, meta-learning, neuromorphic computing, and neuroscience, which we hope to see explored.

\bibliography{references}

\begin{thebibliography}{73}
\providecommand{\natexlab}[1]{#1}
\providecommand{\url}[1]{\texttt{#1}}
\expandafter\ifx\csname urlstyle\endcsname\relax
  \providecommand{\doi}[1]{doi: #1}\else
  \providecommand{\doi}{doi: \begingroup \urlstyle{rm}\Url}\fi

\bibitem[Abraham \& Bear(1996)Abraham and Bear]{abraham1996metaplasticity}
Abraham, W.~C. and Bear, M.~F.
\newblock Metaplasticity: the plasticity of synaptic plasticity.
\newblock \emph{Trends in neurosciences}, 19\penalty0 (4):\penalty0 126--130,
  1996.

\bibitem[Agarwal et~al.(2021)Agarwal, Schwarzer, Castro, Courville, and
  Bellemare]{agarwal2021precipice}
Agarwal, R., Schwarzer, M., Castro, P.~S., Courville, A.~C., and Bellemare,
  M.~G.
\newblock Deep reinforcement learning at the edge of the statistical precipice.
\newblock \emph{CoRR}, abs/2108.13264, 2021.
\newblock URL \url{https://arxiv.org/abs/2108.13264}.

\bibitem[Ba et~al.(2016)Ba, Hinton, Mnih, Leibo, and Ionescu]{ba2016using}
Ba, J., Hinton, G.~E., Mnih, V., Leibo, J.~Z., and Ionescu, C.
\newblock Using fast weights to attend to the recent past.
\newblock \emph{Advances in Neural Information Processing Systems},
  29:\penalty0 4331--4339, 2016.

\bibitem[Banino et~al.(2021)Banino, Balaguer, and
  Blundell]{banino2021pondernet}
Banino, A., Balaguer, J., and Blundell, C.
\newblock Pondernet: Learning to ponder.
\newblock \emph{arXiv preprint arXiv:2107.05407}, 2021.

\bibitem[Beaulieu et~al.(2020)Beaulieu, Frati, Miconi, Lehman, Stanley, Clune,
  and Cheney]{beaulieu2020learning}
Beaulieu, S., Frati, L., Miconi, T., Lehman, J., Stanley, K.~O., Clune, J., and
  Cheney, N.
\newblock Learning to continually learn.
\newblock \emph{arXiv preprint arXiv:2002.09571}, 2020.

\bibitem[{Bellec \& Salaj} et~al.(2018){Bellec \& Salaj}, Subramoney,
  Legenstein, and Maass]{Bellec2018}
{Bellec \& Salaj}, Subramoney, A., Legenstein, R., and Maass, W.
\newblock Long short-term memory and learning-to-learn in networks of spiking
  neurons.
\newblock In Bengio, S., Wallach, H., Larochelle, H., Grauman, K.,
  Cesa-Bianchi, N., and Garnett, R. (eds.), \emph{Advances in Neural
  Information Processing Systems}, volume~31. Curran Associates, Inc., 2018.
\newblock URL
  \url{https://proceedings.neurips.cc/paper/2018/file/c203d8a151612acf12457e4d67635a95-Paper.pdf}.

\bibitem[Bellemare et~al.(2013)Bellemare, Naddaf, Veness, and
  Bowling]{bellemare2013arcade}
Bellemare, M.~G., Naddaf, Y., Veness, J., and Bowling, M.
\newblock The arcade learning environment: An evaluation platform for general
  agents.
\newblock \emph{J. Artif. Intell. Res.}, 47:\penalty0 253--279, 2013.
\newblock \doi{10.1613/jair.3912}.
\newblock URL \url{https://doi.org/10.1613/jair.3912}.

\bibitem[Bengio et~al.(1990)Bengio, Bengio, and Cloutier]{bengio1990learning}
Bengio, Y., Bengio, S., and Cloutier, J.
\newblock \emph{Learning a synaptic learning rule}.
\newblock Citeseer, 1990.

\bibitem[Chamberlain et~al.(2008)Chamberlain, Yang, and
  Jones]{chamberlain2008role}
Chamberlain, S.~E., Yang, J., and Jones, R.~S.
\newblock The role of nmda receptor subtypes in short-term plasticity in the
  rat entorhinal cortex.
\newblock \emph{Neural plasticity}, 2008, 2008.

\bibitem[Euler(1794)]{euler1794}
Euler, L.
\newblock \emph{Institutiones calculi integralis}.
\newblock Academia Imperialis Scientiarum, 1794.

\bibitem[Feldman(1982)]{feldman1982dynamic}
Feldman, J.~A.
\newblock Dynamic connections in neural networks.
\newblock \emph{Biological cybernetics}, 46\penalty0 (1):\penalty0 27--39,
  1982.

\bibitem[Fiebig \& Lansner(2017)Fiebig and Lansner]{fiebig2017spiking}
Fiebig, F. and Lansner, A.
\newblock A spiking working memory model based on hebbian short-term
  potentiation.
\newblock \emph{Journal of Neuroscience}, 37\penalty0 (1):\penalty0 83--96,
  2017.

\bibitem[Flennerhag et~al.(2019)Flennerhag, Rusu, Pascanu, Visin, Yin, and
  Hadsell]{flennerhag2019meta}
Flennerhag, S., Rusu, A.~A., Pascanu, R., Visin, F., Yin, H., and Hadsell, R.
\newblock Meta-learning with warped gradient descent.
\newblock \emph{arXiv preprint arXiv:1909.00025}, 2019.

\bibitem[Gerstner et~al.(2018)Gerstner, Lehmann, Liakoni, Corneil, and
  Brea]{gerstner2018eligibility}
Gerstner, W., Lehmann, M., Liakoni, V., Corneil, D., and Brea, J.
\newblock Eligibility traces and plasticity on behavioral time scales:
  experimental support of neohebbian three-factor learning rules.
\newblock \emph{Frontiers in neural circuits}, 12:\penalty0 53, 2018.

\bibitem[Hausknecht \& Stone(2015)Hausknecht and Stone]{hausknecht2015deep}
Hausknecht, M.~J. and Stone, P.
\newblock Deep recurrent q-learning for partially observable mdps.
\newblock In \emph{2015 {AAAI} Fall Symposia, Arlington, Virginia, USA,
  November 12-14, 2015}, pp.\  29--37. {AAAI} Press, 2015.
\newblock URL
  \url{http://www.aaai.org/ocs/index.php/FSS/FSS15/paper/view/11673}.

\bibitem[Hebb(1949)]{hebb1949organisation}
Hebb, D.~O.
\newblock \emph{The organisation of behaviour: a neuropsychological theory}.
\newblock John Wiley \& Sons, Inc., New York, 1949.

\bibitem[Hinton(2017)]{hinton2017using}
Hinton, G.
\newblock Using fast weights to store temporary memories, 2017.
\newblock URL \url{https://www.youtube.com/watch?v=GLmptInTNSw}.

\bibitem[Hinton \& Plaut(1987)Hinton and Plaut]{hinton1987using}
Hinton, G.~E. and Plaut, D.~C.
\newblock Using fast weights to deblur old memories.
\newblock In \emph{Proceedings of the 9th Annual Conference of the Cognitive
  Science Society}, pp.\  177--186, 1987.

\bibitem[Hochreiter \& Schmidhuber(1997)Hochreiter and
  Schmidhuber]{hochreiter1997long}
Hochreiter, S. and Schmidhuber, J.
\newblock Long short-term memory.
\newblock \emph{Neural computation}, 9\penalty0 (8):\penalty0 1735--1780, 1997.

\bibitem[Hochreiter et~al.(2001)Hochreiter, Younger, and
  Conwell]{Hochreiter01learningto}
Hochreiter, S., Younger, A.~S., and Conwell, P.~R.
\newblock Learning to learn using gradient descent.
\newblock In \emph{IN LECTURE NOTES ON COMP. SCI. 2130, PROC. INTL. CONF. ON
  ARTI NEURAL NETWORKS (ICANN-2001}, pp.\  87--94. Springer, 2001.

\bibitem[Hospedales et~al.(2020)Hospedales, Antoniou, Micaelli, and
  Storkey]{hospedales2020meta}
Hospedales, T., Antoniou, A., Micaelli, P., and Storkey, A.
\newblock Meta-learning in neural networks: A survey.
\newblock \emph{arXiv preprint arXiv:2004.05439}, 2020.

\bibitem[Illing et~al.(2021)Illing, Ventura, Bellec, and
  Gerstner]{illing2021local}
Illing, B., Ventura, J., Bellec, G., and Gerstner, W.
\newblock Local plasticity rules can learn deep representations using
  self-supervised contrastive predictions.
\newblock \emph{Advances in Neural Information Processing Systems}, 34, 2021.

\bibitem[Indiveri(2021)]{indiveri2021introducing}
Indiveri, G.
\newblock Introducing ‘neuromorphic computing and engineering’.
\newblock \emph{Neuromorphic Computing and Engineering}, 1\penalty0
  (1):\penalty0 010401, 2021.

\bibitem[Indiveri \& Horiuchi(2011)Indiveri and
  Horiuchi]{indiveri2011frontiers}
Indiveri, G. and Horiuchi, T.~K.
\newblock Frontiers in neuromorphic engineering, 2011.

\bibitem[Javed \& White(2019)Javed and White]{javed2019meta}
Javed, K. and White, M.
\newblock Meta-learning representations for continual learning.
\newblock In Wallach, H., Larochelle, H., Beygelzimer, A., d\textquotesingle
  Alch\'{e}-Buc, F., Fox, E., and Garnett, R. (eds.), \emph{Advances in Neural
  Information Processing Systems}, volume~32. Curran Associates, Inc., 2019.
\newblock URL
  \url{https://proceedings.neurips.cc/paper/2019/file/f4dd765c12f2ef67f98f3558c282a9cd-Paper.pdf}.

\bibitem[Kendall et~al.(1989)]{kendall1989introduction}
Kendall, E.~A. et~al.
\newblock An introduction to numerical analysis.
\newblock \emph{John Wiley and Sons Inc., New York, USA}, pp.\  37--45, 1989.

\bibitem[Le et~al.(2020)Le, Tran, and Venkatesh]{le2020self}
Le, H., Tran, T., and Venkatesh, S.
\newblock Self-attentive associative memory.
\newblock In \emph{International Conference on Machine Learning}, pp.\
  5682--5691. PMLR, 2020.

\bibitem[LeCun et~al.(1998)LeCun, Bottou, Bengio, and
  Haffner]{lecun1998gradient}
LeCun, Y., Bottou, L., Bengio, Y., and Haffner, P.
\newblock Gradient-based learning applied to document recognition.
\newblock \emph{Proceedings of the IEEE}, 86\penalty0 (11):\penalty0
  2278--2324, 1998.

\bibitem[Leng et~al.(2018)Leng, Martel, Breitwieser, Bytschok, Senn, Schemmel,
  Meier, and Petrovici]{leng2018spiking}
Leng, L., Martel, R., Breitwieser, O., Bytschok, I., Senn, W., Schemmel, J.,
  Meier, K., and Petrovici, M.~A.
\newblock Spiking neurons with short-term synaptic plasticity form superior
  generative networks.
\newblock \emph{Scientific reports}, 8\penalty0 (1):\penalty0 1--11, 2018.

\bibitem[Liang et~al.(2018)Liang, Liaw, Nishihara, Moritz, Fox, Goldberg,
  Gonzalez, Jordan, and Stoica]{dy2018rllib}
Liang, E., Liaw, R., Nishihara, R., Moritz, P., Fox, R., Goldberg, K.,
  Gonzalez, J., Jordan, M., and Stoica, I.
\newblock {RL}lib: Abstractions for distributed reinforcement learning.
\newblock In Dy, J. and Krause, A. (eds.), \emph{Proceedings of the 35th
  International Conference on Machine Learning}, volume~80 of \emph{Proceedings
  of Machine Learning Research}, pp.\  3053--3062. PMLR, 10--15 Jul 2018.
\newblock URL \url{https://proceedings.mlr.press/v80/liang18b.html}.

\bibitem[Limbacher \& Legenstein(2020)Limbacher and
  Legenstein]{limbacher2020hmem}
Limbacher, T. and Legenstein, R.
\newblock H-mem: Harnessing synaptic plasticity with hebbian memory networks.
\newblock In Larochelle, H., Ranzato, M., Hadsell, R., Balcan, M., and Lin, H.
  (eds.), \emph{Advances in Neural Information Processing Systems 33: Annual
  Conference on Neural Information Processing Systems 2020, NeurIPS 2020,
  December 6-12, 2020, virtual}, 2020.
\newblock URL
  \url{https://proceedings.neurips.cc/paper/2020/hash/f6876a9f998f6472cc26708e27444456-Abstract.html}.

\bibitem[L{\"o}we et~al.(2019)L{\"o}we, O'Connor, and Veeling]{lowe2019putting}
L{\"o}we, S., O'Connor, P., and Veeling, B.~S.
\newblock Putting an end to end-to-end: Gradient-isolated learning of
  representations.
\newblock \emph{arXiv preprint arXiv:1905.11786}, 2019.

\bibitem[Maass(1997)]{maass1997networks}
Maass, W.
\newblock Networks of spiking neurons: the third generation of neural network
  models.
\newblock \emph{Neural networks}, 10\penalty0 (9):\penalty0 1659--1671, 1997.

\bibitem[Miconi(2021)]{miconi2021learning}
Miconi, T.
\newblock Learning to acquire novel cognitive tasks with evolution, plasticity
  and meta-meta-learning.
\newblock \emph{arXiv preprint arXiv:2112.08588}, 2021.

\bibitem[Miconi et~al.(2018)Miconi, Stanley, and
  Clune]{miconi2018differentiable}
Miconi, T., Stanley, K., and Clune, J.
\newblock Differentiable plasticity: training plastic neural networks with
  backpropagation.
\newblock In \emph{International Conference on Machine Learning}, pp.\
  3559--3568. PMLR, 2018.

\bibitem[Miconi et~al.(2019)Miconi, Rawal, Clune, and
  Stanley]{miconi2018backpropamine}
Miconi, T., Rawal, A., Clune, J., and Stanley, K.~O.
\newblock Backpropamine: training self-modifying neural networks with
  differentiable neuromodulated plasticity.
\newblock In \emph{International Conference on Learning Representations}, 2019.
\newblock URL \url{https://openreview.net/forum?id=r1lrAiA5Ym}.

\bibitem[Miconi et~al.(2020)Miconi, Rawal, Clune, and
  Stanley]{miconi2020backpropamine}
Miconi, T., Rawal, A., Clune, J., and Stanley, K.~O.
\newblock Backpropamine: training self-modifying neural networks with
  differentiable neuromodulated plasticity.
\newblock \emph{arXiv preprint arXiv:2002.10585}, 2020.

\bibitem[Millidge et~al.(2020)Millidge, Tschantz, and
  Buckley]{millidge2020predictive}
Millidge, B., Tschantz, A., and Buckley, C.~L.
\newblock Predictive coding approximates backprop along arbitrary computation
  graphs.
\newblock \emph{arXiv preprint arXiv:2006.04182}, 2020.

\bibitem[Mnih et~al.(2016)Mnih, Badia, Mirza, Graves, Lillicrap, Harley,
  Silver, and Kavukcuoglu]{mnih2016asynchronous}
Mnih, V., Badia, A.~P., Mirza, M., Graves, A., Lillicrap, T.~P., Harley, T.,
  Silver, D., and Kavukcuoglu, K.
\newblock Asynchronous methods for deep reinforcement learning.
\newblock In Balcan, M. and Weinberger, K.~Q. (eds.), \emph{Proceedings of the
  33nd International Conference on Machine Learning, {ICML} 2016, New York
  City, NY, USA, June 19-24, 2016}, volume~48 of \emph{{JMLR} Workshop and
  Conference Proceedings}, pp.\  1928--1937. JMLR.org, 2016.
\newblock URL \url{http://proceedings.mlr.press/v48/mniha16.html}.

\bibitem[Mongillo et~al.(2008)Mongillo, Barak, and
  Tsodyks]{mongillo2008synaptic}
Mongillo, G., Barak, O., and Tsodyks, M.
\newblock Synaptic theory of working memory.
\newblock \emph{Science}, 319\penalty0 (5869):\penalty0 1543--1546, 2008.

\bibitem[Moraitis et~al.(2018{\natexlab{a}})Moraitis, Sebastian, and
  Eleftheriou]{moraitis2018role}
Moraitis, T., Sebastian, A., and Eleftheriou, E.
\newblock The role of short-term plasticity in neuromorphic learning: learning
  from the timing of rate-varying events with fatiguing spike-timing-dependent
  plasticity.
\newblock \emph{IEEE Nanotechnology Magazine}, 12\penalty0 (3):\penalty0
  45--53, 2018{\natexlab{a}}.

\bibitem[Moraitis et~al.(2018{\natexlab{b}})Moraitis, Sebastian, and
  Eleftheriou]{moraitis2018spiking}
Moraitis, T., Sebastian, A., and Eleftheriou, E.
\newblock Spiking neural networks enable two-dimensional neurons and
  unsupervised multi-timescale learning.
\newblock In \emph{2018 International Joint Conference on Neural Networks
  (IJCNN)}, pp.\  1--8. IEEE, 2018{\natexlab{b}}.

\bibitem[Moraitis et~al.(2020)Moraitis, Sebastian, and
  Eleftheriou]{moraitis2020optimality}
Moraitis, T., Sebastian, A., and Eleftheriou, E.
\newblock Optimality of short-term synaptic plasticity in modelling certain
  dynamic environments.
\newblock \emph{arXiv preprint arXiv:2009.06808}, 2020.

\bibitem[Moraitis et~al.(2021)Moraitis, Toichkin, Chua, and
  Guo]{moraitis2021softhebb}
Moraitis, T., Toichkin, D., Chua, Y., and Guo, Q.
\newblock Softhebb: Bayesian inference in unsupervised hebbian soft
  winner-take-all networks, 2021.

\bibitem[Nessler et~al.(2009)Nessler, Pfeiffer, and Maass]{nessler2009stdp}
Nessler, B., Pfeiffer, M., and Maass, W.
\newblock Stdp enables spiking neurons to detect hidden causes of their inputs.
\newblock \emph{Advances in neural information processing systems},
  22:\penalty0 1357--1365, 2009.

\bibitem[Pogodin \& Latham(2020)Pogodin and Latham]{pogodin2020}
Pogodin, R. and Latham, P.
\newblock Kernelized information bottleneck leads to biologically plausible
  3-factor hebbian learning in deep networks.
\newblock In Larochelle, H., Ranzato, M., Hadsell, R., Balcan, M.~F., and Lin,
  H. (eds.), \emph{Advances in Neural Information Processing Systems},
  volume~33, pp.\  7296--7307. Curran Associates, Inc., 2020.
\newblock URL
  \url{https://proceedings.neurips.cc/paper/2020/file/517f24c02e620d5a4dac1db388664a63-Paper.pdf}.

\bibitem[Ponulak \& Kasinski(2011)Ponulak and
  Kasinski]{ponulak2011introduction}
Ponulak, F. and Kasinski, A.
\newblock Introduction to spiking neural networks: Information processing,
  learning and applications.
\newblock \emph{Acta neurobiologiae experimentalis}, 71\penalty0 (4):\penalty0
  409--433, 2011.

\bibitem[Ren et~al.(2018)Ren, Liao, Fetaya, and Zemel]{ren2018incremental}
Ren, M., Liao, R., Fetaya, E., and Zemel, R.~S.
\newblock Incremental few-shot learning with attention attractor networks.
\newblock \emph{arXiv preprint arXiv:1810.07218}, 2018.

\bibitem[Rosenbaum et~al.(2012)Rosenbaum, Rubin, and
  Doiron]{rosenbaum2012short}
Rosenbaum, R., Rubin, J., and Doiron, B.
\newblock Short term synaptic depression imposes a frequency dependent filter
  on synaptic information transfer.
\newblock \emph{PLoS computational biology}, 8\penalty0 (6):\penalty0 e1002557,
  2012.

\bibitem[Salimans \& Kingma(2016)Salimans and Kingma]{salimans2016weight}
Salimans, T. and Kingma, D.~P.
\newblock Weight normalization: A simple reparameterization to accelerate
  training of deep neural networks.
\newblock \emph{Advances in neural information processing systems}, 29, 2016.

\bibitem[Sarwat et~al.(2022{\natexlab{a}})Sarwat, Kersting, Moraitis,
  Jonnalagadda, and Sebastian]{sarwat2022phase}
Sarwat, S.~G., Kersting, B., Moraitis, T., Jonnalagadda, V.~P., and Sebastian,
  A.
\newblock Phase-change memtransistive synapses for mixed-plasticity neural
  computations.
\newblock \emph{Nature Nanotechnology}, 17\penalty0 (5):\penalty0 507--513,
  2022{\natexlab{a}}.

\bibitem[Sarwat et~al.(2022{\natexlab{b}})Sarwat, Moraitis, Wright, and
  Bhaskaran]{sarwat2022chalcogenide}
Sarwat, S.~G., Moraitis, T., Wright, C.~D., and Bhaskaran, H.
\newblock Chalcogenide optomemristors for multi-factor neuromorphic
  computation.
\newblock \emph{Nature communications}, 13\penalty0 (1):\penalty0 1--9,
  2022{\natexlab{b}}.

\bibitem[Scellier \& Bengio(2019)Scellier and Bengio]{scellier2019equivalence}
Scellier, B. and Bengio, Y.
\newblock Equivalence of equilibrium propagation and recurrent backpropagation.
\newblock \emph{Neural computation}, 31\penalty0 (2):\penalty0 312--329, 2019.

\bibitem[Schlag \& Schmidhuber(2017)Schlag and Schmidhuber]{schlag2017gated}
Schlag, I. and Schmidhuber, J.
\newblock Gated fast weights for on-the-fly neural program generation.
\newblock In \emph{NIPS Metalearning Workshop}, 2017.

\bibitem[Schlag et~al.(2021)Schlag, Irie, and Schmidhuber]{schlag2021linear}
Schlag, I., Irie, K., and Schmidhuber, J.
\newblock Linear transformers are secretly fast weight programmers.
\newblock In \emph{International Conference on Machine Learning}, pp.\
  9355--9366. PMLR, 2021.

\bibitem[Schmidhuber(1992)]{schmidhuber1992learning}
Schmidhuber, J.
\newblock Learning to control fast-weight memories: An alternative to dynamic
  recurrent networks.
\newblock \emph{Neural Computation}, 4\penalty0 (1):\penalty0 131--139, 1992.

\bibitem[Schmidhuber(1993)]{schmidhuber1993reducing}
Schmidhuber, J.
\newblock Reducing the ratio between learning complexity and number of time
  varying variables in fully recurrent nets.
\newblock In \emph{International Conference on Artificial Neural Networks},
  pp.\  460--463. Springer, 1993.

\bibitem[Schmidhuber et~al.(1996)Schmidhuber, Zhao, and
  Wiering]{schmidhuber1996simple}
Schmidhuber, J., Zhao, J., and Wiering, M.
\newblock Simple principles of metalearning.
\newblock \emph{Technical report IDSIA}, 69:\penalty0 1--23, 1996.

\bibitem[Schulman et~al.(2017)Schulman, Wolski, Dhariwal, Radford, and
  Klimov]{schulman2017proximal}
Schulman, J., Wolski, F., Dhariwal, P., Radford, A., and Klimov, O.
\newblock Proximal policy optimization algorithms.
\newblock \emph{CoRR}, abs/1707.06347, 2017.
\newblock URL \url{http://arxiv.org/abs/1707.06347}.

\bibitem[Schwarzschild et~al.(2021{\natexlab{a}})Schwarzschild, Borgnia, Gupta,
  Huang, Vishkin, Goldblum, and Goldstein]{schwarzschild2021can}
Schwarzschild, A., Borgnia, E., Gupta, A., Huang, F., Vishkin, U., Goldblum,
  M., and Goldstein, T.
\newblock Can you learn an algorithm? generalizing from easy to hard problems
  with recurrent networks.
\newblock \emph{Advances in Neural Information Processing Systems}, 34,
  2021{\natexlab{a}}.

\bibitem[Schwarzschild et~al.(2021{\natexlab{b}})Schwarzschild, Gupta, Ghiasi,
  Goldblum, and Goldstein]{schwarzschild2021uncanny}
Schwarzschild, A., Gupta, A., Ghiasi, A., Goldblum, M., and Goldstein, T.
\newblock The uncanny similarity of recurrence and depth.
\newblock \emph{arXiv preprint arXiv:2102.11011}, 2021{\natexlab{b}}.

\bibitem[Sinitsin et~al.(2020)Sinitsin, Plokhotnyuk, Pyrkin, Popov, and
  Babenko]{Sinitsin2020Editable}
Sinitsin, A., Plokhotnyuk, V., Pyrkin, D., Popov, S., and Babenko, A.
\newblock Editable neural networks.
\newblock In \emph{International Conference on Learning Representations}, 2020.
\newblock URL \url{https://openreview.net/forum?id=HJedXaEtvS}.

\bibitem[Soltoggio et~al.(2018)Soltoggio, Stanley, and Risi]{soltoggio2018born}
Soltoggio, A., Stanley, K.~O., and Risi, S.
\newblock Born to learn: the inspiration, progress, and future of evolved
  plastic artificial neural networks.
\newblock \emph{Neural Networks}, 108:\penalty0 48--67, 2018.

\bibitem[Szatm{\'a}ry \& Izhikevich(2010)Szatm{\'a}ry and
  Izhikevich]{szatmary2010spike}
Szatm{\'a}ry, B. and Izhikevich, E.~M.
\newblock Spike-timing theory of working memory.
\newblock \emph{PLoS computational biology}, 6\penalty0 (8):\penalty0 e1000879,
  2010.

\bibitem[Thrun \& Pratt(1998)Thrun and Pratt]{thrun1998learning}
Thrun, S. and Pratt, L.
\newblock Learning to learn: Introduction and overview.
\newblock In \emph{Learning to learn}, pp.\  3--17. Springer, 1998.

\bibitem[Tieleman \& Hinton(2009)Tieleman and Hinton]{tieleman2009using}
Tieleman, T. and Hinton, G.
\newblock Using fast weights to improve persistent contrastive divergence.
\newblock In \emph{Proceedings of the 26th annual international conference on
  machine learning}, pp.\  1033--1040, 2009.

\bibitem[Todorov et~al.(2012)Todorov, Erez, and Tassa]{todorov2012mujoco}
Todorov, E., Erez, T., and Tassa, Y.
\newblock Mujoco: A physics engine for model-based control.
\newblock In \emph{2012 IEEE/RSJ International Conference on Intelligent Robots
  and Systems}, pp.\  5026--5033. IEEE, 2012.

\bibitem[Tsodyks \& Markram(1997)Tsodyks and Markram]{tsodyks1997neural}
Tsodyks, M.~V. and Markram, H.
\newblock The neural code between neocortical pyramidal neurons depends on
  neurotransmitter release probability.
\newblock \emph{Proceedings of the national academy of sciences}, 94\penalty0
  (2):\penalty0 719--723, 1997.

\bibitem[Tyulmankov et~al.(2022)Tyulmankov, Yang, and
  Abbott]{tyulmankov2022meta}
Tyulmankov, D., Yang, G.~R., and Abbott, L.
\newblock Meta-learning synaptic plasticity and memory addressing for continual
  familiarity detection.
\newblock \emph{Neuron}, 110\penalty0 (3):\penalty0 544--557, 2022.

\bibitem[Vuorio et~al.(2018)Vuorio, Cho, Kim, and Kim]{vuorio2018meta}
Vuorio, R., Cho, D.-Y., Kim, D., and Kim, J.
\newblock Meta continual learning.
\newblock \emph{arXiv preprint arXiv:1806.06928}, 2018.

\bibitem[Wang et~al.(2016)Wang, Kurth-Nelson, Tirumala, Soyer, Leibo, Munos,
  Blundell, Kumaran, and Botvinick]{wang2016learning}
Wang, J.~X., Kurth-Nelson, Z., Tirumala, D., Soyer, H., Leibo, J.~Z., Munos,
  R., Blundell, C., Kumaran, D., and Botvinick, M.
\newblock Learning to reinforcement learn.
\newblock \emph{arXiv preprint arXiv:1611.05763}, 2016.

\bibitem[Wo{\'z}niak et~al.(2020)Wo{\'z}niak, Pantazi, Bohnstingl, and
  Eleftheriou]{wozniak2020deep}
Wo{\'z}niak, S., Pantazi, A., Bohnstingl, T., and Eleftheriou, E.
\newblock Deep learning incorporating biologically inspired neural dynamics and
  in-memory computing.
\newblock \emph{Nature Machine Intelligence}, 2\penalty0 (6):\penalty0
  325--336, 2020.

\bibitem[Zucker(1989)]{zucker1989short}
Zucker, R.~S.
\newblock Short-term synaptic plasticity.
\newblock \emph{Annual review of neuroscience}, 12\penalty0 (1):\penalty0
  13--31, 1989.

\end{thebibliography}
\bibliographystyle{icml2022}

\newpage
\appendix
\onecolumn
\section{Appendix}
\label{sec:Appendix}
\subsection{The meta-learning algorithm}
\begin{algorithm}[h]
	\caption{STPN learning to learn and forget in a supervised meta-learning setting}
	\label{alg:example}
	\begin{algorithmic}
		\STATE Initialize weights and meta-parameters $\boldsymbol{\Theta} = [\boldsymbol{W};\boldsymbol{\lambda};\boldsymbol{\gamma}]$
		\STATE \(\triangleright\) Outer-loop learning. E.g.\  BPTT.
		\WHILE {training}
		\STATE \(\triangleright\) Sample input sequence.
		\STATE $\mathbfcal{S} = [\boldsymbol{x}^{(0)}, \ldots , \boldsymbol{x}^{(T)}] \sim \mathcal{D}$
		\STATE \(\triangleright\) Initialize short-term component.
		\STATE $\boldsymbol{F}^{(0)} = \mathbf{0}$
		\STATE \(\triangleright\) Inner-loop learning via Hebbian STP $\equiv$ Recurrent model receiving an input sequence.
		\FOR{$t=1$ {\bfseries to} $T$ }
		\STATE \(\triangleright\) Obtain total efficacies.
		\STATE $\boldsymbol{G}^{(t)}=\boldsymbol{W}+\boldsymbol{F}^{(t)}$
		\textcolor{gray}{
			\STATE \(\triangleright\) Normalize total efficacies and their short-term components (optional but practically impactful).
			\STATE $\boldsymbol{\hat{G}}^{(t)}=\boldsymbol{G}^{(t)} / \lVert \boldsymbol{G}^{(t)} \rVert$
			\STATE $\boldsymbol{\hat{F}}^{(t)}=\boldsymbol{F}^{(t)} / \lVert \boldsymbol{G}^{(t)} \rVert$}
		
		\STATE \(\triangleright\) Forward pass.
		\STATE $\boldsymbol{h}^{(t)}=\sigma(\boldsymbol{\hat{G}}^{(t)} \boldsymbol{x}^{(t)})$
		
		\STATE \(\triangleright\) Hebbian STP update $\equiv$ Synaptic-recurrent-state update.
		\STATE $\boldsymbol{F}^{(t+1)}=\boldsymbol{\gamma} \odot (\boldsymbol{x}^{(t)} \otimes \boldsymbol{h}^{(t)})+\boldsymbol{\lambda} \odot \boldsymbol{\hat{F}}^{(t)}$
		\ENDFOR
		
		\STATE \(\triangleright\) Meta-learn via gradient update, e.g.\ BPTT.
		\STATE $\boldsymbol{\Theta} \gets \boldsymbol{\Theta} - \nabla \mathcal{L}$
		\ENDWHILE
	\end{algorithmic}
\end{algorithm}

\subsection{Importance of learning per-synapse STP parameters}
\label{ssec:Appendix:per-synapse-ablation}
\cref{table:per-synapse-vs-uniform-vs-hebbff-art-maze-eval} shows that learning per-synapse STP parameters achieves greater proficiency and efficiency. This happens for both STPNf and STPNr, respectively without and with recurrent connections. Note energy efficiency being lower for purely feed-forward models, as happens for ART, can be expected. This is because recurrent activations are dense, whereas input for these problems is more sparse and lower magnitude on average, which we do not correct for in these measurements. All models had their hidden size chosen to have similar number of parameters.

\begin{table*}[!h]
	\caption{Inference-time proficiency and energy efficiency of per-synapse and uniform STP models on ART and Maze Exploration}
	\begin{center}
		\begin{tabular}{lllll}
			\toprule
			{} &  \multicolumn{2}{l}{ART} &  \multicolumn{2}{l}{Maze} \\
			\midrule
			Model &  Test accuracy & Power consumption & Reward & Power consumption \\
			\midrule
			STPNr per-synapse STP (Ours) & \textbf{99.99  $\pm$ 0.01} &       3.4  $\pm$ 0.4 & \textbf{115.7  $\pm$ 1.6} &      \textbf{80.2  $\pm$ 2.4} \\
			STPNf per-synapse STP (Ours) & 89.40  $\pm$ 5.14 &       \textbf{1.2  $\pm$ 0.1} & 112.9  $\pm$ 1.5 &     154.8  $\pm$ 6.3\\
			STPNr uniform STP (Ours) & 96.89  $\pm$ 6.21 &       4.0  $\pm$ 0.3 & 74.0  $\pm$ 1.4 &     150.2  $\pm$ 5.5 \\
			STPNf uniform STP (Ours) & 60.21  $\pm$ 0.52 &       2.6  $\pm$ 0.1 & 85.2  $\pm$ 1.6 &    355.6  $\pm$ 12.9\\
			HebbFF (Tyulmankov et al., 2022) & 10.62  $\pm$ 0.02 &       7.9  $\pm$ 1.8  &  39.9  $\pm$ 1.3 &     385.9  $\pm$ 8.9\\
			\bottomrule
		\end{tabular}
	\end{center}
	\label{table:per-synapse-vs-uniform-vs-hebbff-art-maze-eval}
\end{table*}


ART and Maze Exploration tasks were presented in other works presenting different networks with plasticity, so they serve as solid baselines for STPN without an explicit task bias introduced by our work. Nonetheless, HebbFF's performance on these two tasks is far below the structurally equivalent STPNf. Therefore, we decide to test the importance of recurrence and per-synapse STP in the Continual Familiarity Detection Task presented in \cite{tyulmankov2022meta}. In this task, at each timestep the network is presented with a binary vector whose elements are $1$ or $-1$. With probability $1 - p$ , this vector is randomly sampled. Otherwise, with probability $p$, the vector provided to the network $R$ timesteps ago is used as input again. One exception is that, if an already repeated input was sampled to be repeated again, a new vector is generated instead. The network outputs a single bit at every timestep, representing whether it predicts the vector presented is repeated or not.

We report results on the same three variations of this task presented in \citet{tyulmankov2022meta}, running 5 seeds for each variation of the task. In the first variation, a single dataset of random vectors is used for all iterations of training ('dataset' mode). In the following two, a new set of vectors is generated at each training iteration ('infinite' mode). The distance between repeated vectors ($R$) is also varied from 3 to 6 in the last two experiments. We use T=5000 vectors per iteration (as the experiment logs in the public repository for \citet{tyulmankov2022meta}), and train for the same number of epochs as the released models \cite{tyulmankov2022meta} (although it is not clear whether these were originally trained until convergence). All other training configurations are exactly the same as found in their publicly available code. We choose the hidden size of all other variants as to have similar number of parameters to HebbFF. Given HebbFF has a much higher memory consumption for the same number of trainable parameters than per-synapse STPNr, the STPNr has nearly half the number of hidden units in the Maze exploration for these comparison results as compared to those shown in \cref{fig:proficiency,fig:efficiency}, which explains the lower performance.


\cref{table:per-synapse-vs-uniform-vs-hebbff-contFam} shows the generalization accuracy for the three tasks we described. For the 'dataset' mode, this means accuracy on a validation set. For the 'infinite' mode, accuracy is measured on a newly generated dataset at each iteration, which we label as 'train accuracy' although the network hasn't seen these specific examples before (hence generalization). The results confirm that STPN, especially in its per-synapse variants, performs competitively with HebbFF. Note that we did not tune any of the STPN variants for this task at all, whereas we can HebbFF to be well tuned. STPN is clearly stronger in the infinite mode with R=3, which happens to be the one with the shortest training iterations (1600, vs 3000 for 'dataset' mode and 3500 for the other 'infinite' mode task). Possibly training STPN until convergence would have led to more competitive results, in addition to tuning. Another reason for which HebbFF is better suited for this task is the domain restriction of STP parameters for HebbFF (e.g. only anti-Hebbian and short-term behavior) can be specially suited to this task but not generalize to more complex problems. Furthermore, the plastic update stabilization and initialization we use for STPN might be better suited for more complex problems. This results confirm the use of recurrence and per-synapse STP as a better general approach to configuring STPN.

\begin{table*}[!h]
	\caption{Generalization proficiency of per-synapse and uniform STP models and LSTM in three variants of the Continual Familiarity Detection Task \cite{tyulmankov2022meta}}
	\begin{center}
		\begin{tabular}{llll}
			\toprule
			Model & Dataset mode (R=3) & Infinite mode (R=3) & Infinite mode (R=6)\\
			\midrule
			STPNr per-synapse STP (Ours) &  \textbf{97.53 $\pm$ 0.31} &    \textbf{99.99 $\pm$ 0.02} &    98.98 $\pm$ 0.28 \\
			STPNf per-synapse STP (Ours) & \textbf{98.16 $\pm$ 0.91} &    99.92 $\pm$ 0.04 &    98.72 $\pm$ 0.19\\
			STPNr uniform STP (Ours) & 78.52 $\pm$ 12.82 &    94.13 $\pm$ 7.59 &    90.88 $\pm$ 3.00 \\
			STPNf uniform STP (Ours) & 81.08 $\pm$ 14.27 &    91.95 $\pm$ 9.45 &   91.50 $\pm$ 11.67 \\
			HebbFF (Tyulmankov et al., 2022) & \textbf{98.47 $\pm$ 0.87} &    99.68 $\pm$ 0.09 &    \textbf{99.71 $\pm$ 0.12} \\
			LSTM & 66.75 $\pm$ 0.39 &    78.52 $\pm$ 2.07 &    74.37 $\pm$ 2.90 \\
			\bottomrule
		\end{tabular}
	\end{center}
	\label{table:per-synapse-vs-uniform-vs-hebbff-contFam}
\end{table*}

\begin{figure*}[h!]
	\centering
	\includegraphics[width = \textwidth]{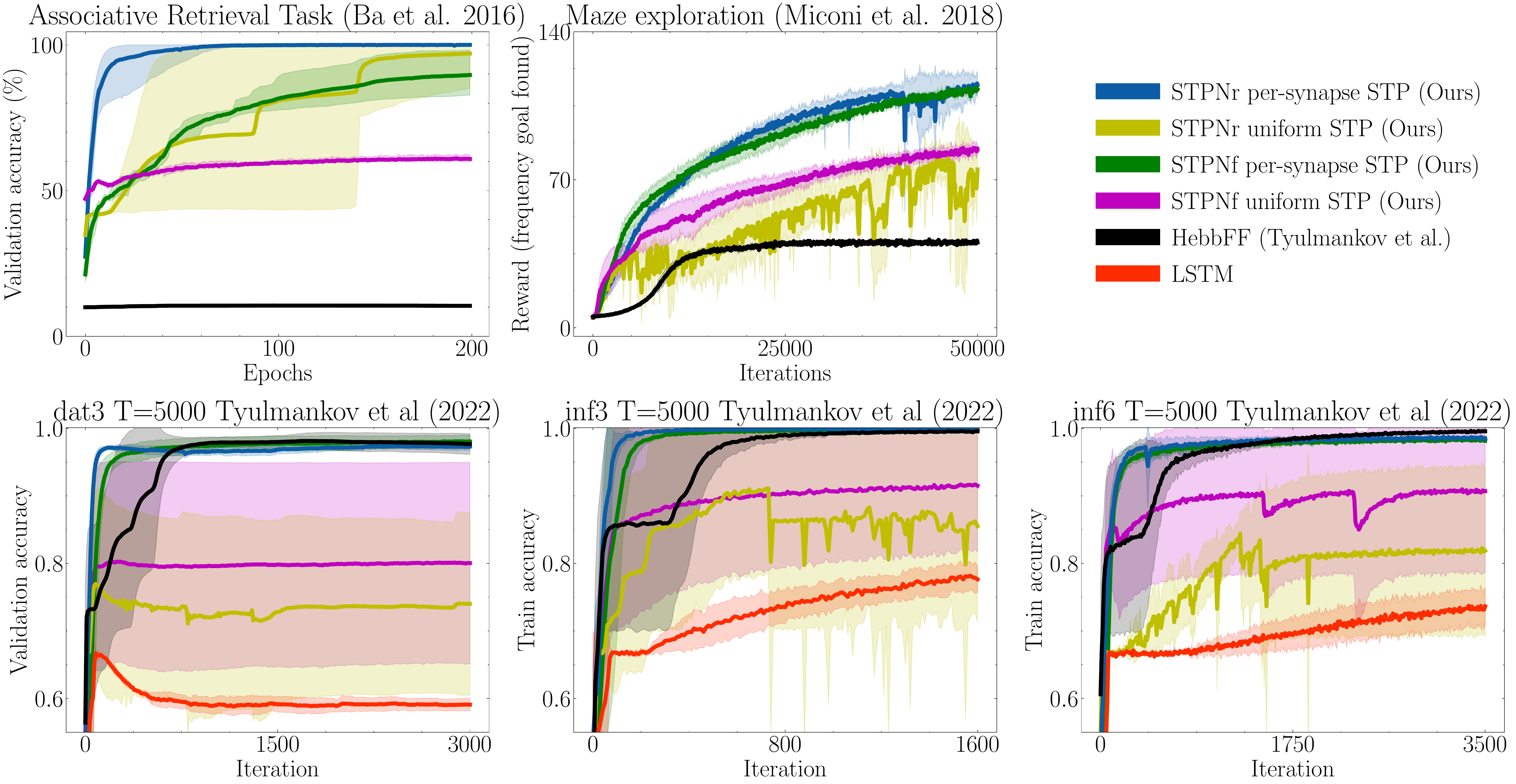}
	\caption{Generalization proficiency curves during training of per-synapse and uniform STP models and LSTM, in ART, Maze Exploration and three variants of the Continual Familiarity Detection Task \cite{tyulmankov2022meta}}
	\label{fig:per-synapse-vs-uniform-vs-hebbff-art-maze-contFam}
\end{figure*}

\subsection{Further results}
For completeness, we present the inference-time equivalents of \cref{fig:proficiency,fig:efficiency} in tabular form, which can be found in \cref{table:proficiency,table:efficiency}. STPN has the added benefit of being able to adapt its synapses after training and hence having a better adaptation during inference, which clearly provides an added extra performance at test time.

\begin{table*}[!h]
	\caption{Accuracy and reward of trained models during inference. Averaged over multiple evaluating runs, reported mean and standard deviation of different random seeds.}
	\begin{center}
		\begin{tabular}{lllll}
			\toprule
			Model & Associative Retrieval & Maze Exploration & Pong & Inverted Pendulum \\
			
			\midrule
			STPN (Ours)                                &                         \textbf{98.55  $\pm$ 2.76} &                    \textbf{132.7  $\pm$ 1.4} &       \textbf{20.7  $\pm$ 0.5} & \textbf{985.0 $\pm$ 15.0}\\
			LSTM                                       &                         47.28  $\pm$ 3.16 &                    119.8  $\pm$ 2.1 &       18.0  $\pm$ 4.6 & 23.2 $\pm$ 10.1	\\
			RNN                                        &                         46.83  $\pm$ 5.56 &                     97.9  $\pm$ 1.9 &                  - & 143.0 $\pm$ 22.4\\
			Modulated Plasticity &                        61.49  $\pm$ 15.79 &                    112.2  $\pm$ 2.0 &                  - & - \\
			Fast Weights             &                         80.87  $\pm$ 1.67 &                                - &                  - & - \\
			MLP             &                          - &                                - &                  - & 701.8 $\pm$ 98.9 \\
			\bottomrule
		\end{tabular}
	\end{center}
	\label{table:proficiency}
\end{table*}

\begin{table*}[!h]
	\caption{Power consumption of trained models during inference. Average over multiple inference runs and timesteps within a run, reported mean and standard deviation over seeds.}
	\begin{center}
		\begin{tabular}{lllll}
			\toprule
			Model & Associative Retrieval & Maze Exploration & Pong & Inverted Pendulum\\
			\midrule
			STPN (Ours)                                &                           \textbf{10.9  $\pm$ 3.8} &                    \textbf{181.1  $\pm$ 2.9} &       \textbf{52.7  $\pm$ 6.4} &  \textbf{7.9 $\pm$ 2.5} \\
			LSTM                                       &                           65.6  $\pm$ 3.4 &                   649.1  $\pm$ 15.3 &     576.8  $\pm$ 33.5 & 758.8 $\pm$ 81.7\\
			RNN                                        &                           43.0  $\pm$ 4.9 &                   330.6  $\pm$ 24.3 &                  -  & 143.0 $\pm$ 22.4 \\
			Modulated plasticity &                           32.0  $\pm$ 7.4 &                   231.6  $\pm$ 23.9 &                  -   & - \\
			Fast Weights             &                          80.6  $\pm$ 69.4 &                                - &                  -  & - \\
			MLP             &                          - &                                - &                  - & 115.8 $\pm$ 33.9 \\
			\bottomrule
		\end{tabular}
	\end{center}
	\label{table:efficiency}
\end{table*}

\subsection{Training methods}
\label{sec:app_trainingmethods}





\paragraph{Stabilization through a type of weight normalization.}Plastic weights can significantly impact the predictions of a network based on recent experience. Although significant changes to the effective weights can increase predictive performance, fast changes of connectionist parameters have the potential to cause a variety of issues during training and evaluation. For instance, \citet{ba2016using} found that large or small norms in hidden vectors can cause the fast weights derived from them to grow or shrink excessively, leading to exploding or vanishing gradients. In their case, using layer normalization of hidden states, and performing a decay of fast weights, successfully controls such instabilities. \citet{miconi2018backpropamine} cites the inherent instability of Hebbian updates as the reason for inclusion of a clipping mechanism for synaptic memories. Alternatively, it also experiments with an implicit normalization (by using Oja's rule instead of Hebb's rule for the plastic weight updates), or a decay term (not-trained, a function of the 'fast learning rate' which is uniform for all synapses); ultimately choosing the clipping due to superior empirical results. 

In our case, layer normalization of hidden inputs alone is not enough to counteract hidden states of varying norm, as input also forms part of the Hebbian update. Additionally, the gradients of STP parameters $\lambda$ and $\gamma$, which are trainable in the STPN, suffer from further instability as these parameters are connected to the computational graph only through the updating of $\boldsymbol{F}$. Relying on decaying $\boldsymbol{F}$ is also not an option, given we do not to restrict $\lambda$ to cause a decay of synaptic memories (as we found restricting it to hinder proficiency). In fact, in some cases, some synapses learn to potentiate their current memories. This adds to the issue of growing memories, especially if the updates to synaptic memories happen to also be purely additive. We experiment with multiple of these techniques, on top of explicit normalization of weights; which ultimately turns out to be the best option in most cases. This weight normalization is performed on the effective weights for each batch element at each timestep, which also normalizes the synaptic memories. Note however we store the un-normalized long-term weights, as these are common to all batch elements and an average norm across the batch defeats the purpose of sequence specific weight adaptation. Additionally, this allows to implicitly learn the relative norm between long-term and short-term weights, given no scaling of the normalized weights is learned as in standard weight normalization \cite{salimans2016weight}. We performed normalization per neuron, although similar performance was observed for global normalization and could be further explored. Additional learnable norm scaling is left for future work. To confirm weight normalization did not provide an inherent training improvement for all models (but only for STPN due to the previously cited reasons), we trained all baselines in ART and Maze with equivalent non-parametric weight normalization schemes, and find neither proficiency or efficiency are improved for other baselines.

\paragraph{Initialization strategy of STP parameters.}
Initialization of STP parameters $\lambda$ and $\gamma$ can also have a significant impact on convergence speed. We tested different random initialization distributions in ART, where we find the best scheme that we use in all other experiments. We find initializing $\gamma_{ij} \sim \mathcal{U}(-\frac{0.001}{\sqrt{h}}, \frac{0.001}{\sqrt{h}})$ is better over purely positive, centered away from zero, or more largely spread values. We hypothesize small initial values of $\gamma$ allow long-term weights to have a greater impact on very early training as short-term weights are very weakly updated. Surprisingly, a much wider and purely positive initialization $\lambda_{ij} \sim \mathcal{U}(0, 1)$ is empirically better. Presumably, decay doesn't play a major role at very early stages of training when $\gamma$ is very close to zero, and in later stages of learning when plastic weights start playing a role, having large variety of dynamics across synapses and neurons allows for richer evolution of synaptic memories.

\subsection{Experimental details}
\label{sec:app_exp}
We run a different number of seeds across experiments, running more when either or both experiment was not highly computationally expensive or more instances were necessary to make clearer conclusions. Specifically, we used 5 (ART and Pong), 3 (Maze) and 2 (Pendulum) seeds respectively.

\subsubsection{Associative Retrieval}
\label{sssec:details-art}
 Compared to the described setup in \citet{ba2016using}, a modification in the experiments carried out is we do not include an embedding or (post-RNN, ) pre-SoftMax fully connected layer, of size 100 each. The motivation behind such a choice is three fold: a) Reducing the network to only a RNN and a SoftMax layer makes the predictive capacity of the Network rely maximally on the RNN, which is the true subject of study b) the second author admits to the lack of need for such embedding \cite{hinton2017using}, and attributes its inclusion to reasons beyond the scope of such paper, and c) when tuning the learning rate we find all models (including non-plastic baselines with small hidden units) easily solve the task if an embedding layer is included, which we hypothesize is caused by such layer taking a major role in prediction.
 
 We tune the learning rate (to 0.001), which is not provided in \citet{ba2016using}. The number of training epochs for the results reported in \cite{ba2016using} is not provided either. We experiment with some values and report results on networks trained for 200 epochs, after which we find networks generally do not improve. Hidden sizes of networks are chosen so that a similar number of parameters are trained, with a focus on low-parameter regime as larger networks all solve the problem (also reported in \citet{ba2016using}). Therefore we choose a hidden size of 20 (the smallest tested size in \citet{ba2016using}) for the Fast weights RNN, and calculate the hidden size of other models as to obtain a similar number of parameters of the entire network. For RNN with fast weights, we use layer normalization, only one iteration of the inner loop,  and $\lambda$ and $\eta$, all as indicated in the Appendix A.1 in \citet{ba2016using}. Although the network structure is different, we don't change the STP parameters ($\lambda$ and $\eta$) as they the same values are used throughout the different tasks in the paper due to alleged robustness provided by layer normalization. Our experiments tuning fast weights RNN confirms this, as they show little difference when exploring other values. Additionally, we use hyperbolic tangent as activation function, for fairer comparison to other networks and as our experiments show a better performance than with ReLU (which only provides purely additive Hebbian factors). Initialization of recurrent (slow) weights as an identity matrix is know to favor longer (orthogonal) storing of hidden states in RNNs \cite{hinton2017using} and hence don't offer a specific advantage to RNN with fast weights vs other RNNs, so the results shown in this paper don't follow such specific initialization for recurrent connections.

\subsubsection{Maze Exploration}
We use the same parameters provided in \cite{miconi2018backpropamine}, and we run the experiments by adapting the open source repository (hence also using the parameters not explicitly mentioned in the article but introduced in the code). We only increase the batch size (number of agents 'acting in parallel') from 16 (in the code, not mentioned in the article) to 512 to maximize computational efficiency of gradient updates. Besides faster learning in a shorter amount of iterations due to more experience per gradient update, we find it can reduce variance of the update as in some episodes initial exploration for the reward fails or comes too late in the episode, hence the agent not being able to exploit learning the location of the reward. Regarding the hidden sizes of the RNNs, we choose a hidden size of 100 for the standard RNN (as provided in the standard \citet{miconi2018backpropamine} parameters), and calculate the hidden size of other models as to obtain a similar number of parameters for the entire network. Additionally, following \citet{miconi2018backpropamine}, value and action branches are non plastic for all models, only fixed and feed forward.

\subsubsection{Atari Pong}
We use RLLib \cite{dy2018rllib} to train and evaluate agents in PongNoFrameskip-v4. Besides the experimental choices described in \cref{ssec:methods:tasks}, it should be noted that we use STPNr with 64 hidden units, and adjust the size of the LSTM to 48 hidden units in order to have a similar number of trainable parameters. We also follow \citet{mnih2016asynchronous} in the preprocessing (dimensionality and color scale) for the game frames, besides frame stacking. We tune rollout length (50), gradient clipping (40), discount factor (0.99) in shorter runs (which both models share in the displayed results); and additionally tune initial learning rate for the final longer runs (0.0007 and 0.0001 respectively), using a linear decay learning rate schedule finishing at $10^{-11}$ at 200 million iterations. Models are trained from the experienced collected by 64 parallel agents.



\subsubsection{MuJoCo Inverted Pendulum}
We use RLLib to train and evaluate agents in InvertedPendulum-v2. We mostly do so with the same parameters reported in \cite{schulman2017proximal}, and for any other parameters not specified there we use the default options provided by RLLib. The only differences are: a) the use of a linearly decaying learning rate schedule, starting from the same original learning rate and finishing at 4 million timesteps at $10^{-11}$; b) since it's not specified in \cite{schulman2017proximal}, we employ a single parallel worker collecting experience in one environment c) the choice of a single layer policy network (besides de action and value branch) of size 64 (this is the only case where models' hidden sizes were not adjusted to obtain a similar number of parameters, as it didn't prove relevant in our experiments).



\subsection{Terminology on timescales}
\label{sec:app_term}
An explicit note about the different timescales across experiments might ease the interpretation of the reported results by the reader, and hence we now provide it. In the supervised learning task, namely ART, an epoch represents a training iteration in which the entire training dataset is seen, and sequence elements describe each of the characters presented to the network in a sequential manner. In reinforcement learning, where an agent interacts with an environment via actions, and receives rewards and observations; we refer to each of these interactions as an episode step or timestep. An episode encompasses a variable (Pong, Inverted Pendulum) or fixed (Maze) number of episode steps or timesteps. Finally, although iteration could also refer to a single timestep, we use such term to indicate one period between gradient updates, which can enclose multiple episodes experienced by multiple agents, depending on the learning algorithm.

\pagebreak
\subsection{STPN mechanics}
\begin{figure}[h!]
	\centering
	\includegraphics[width = .9\textwidth]{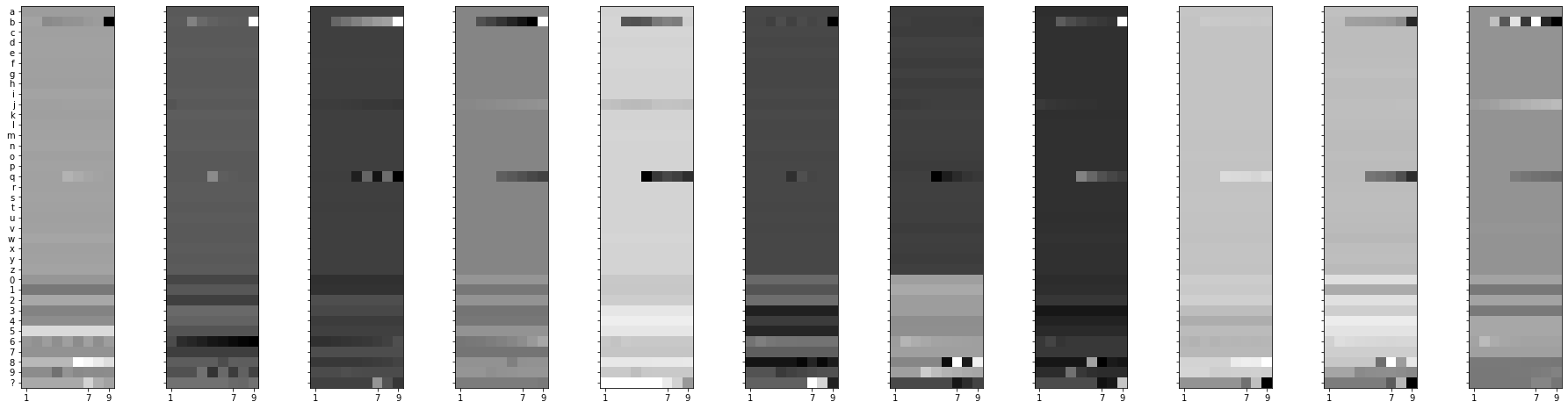}
	\caption{Evolution of effective weights through processing of a sequence in ART. Each separate plot represents a neuron, where the y axis are the input synapses for such neuron, and the x axis represents time, i.e.\ sequence elements. Color represents the magnitude of $G_{ij}$ for each synapse. We can observe the effect of training $\lambda_{ij}$ as different synapses increase or decrease in value at different rates}
	\label{fig:mechanistic_stpn_art}
\end{figure}

\begin{figure}[h!]
	\centering
	\includegraphics[width = .5\textwidth]{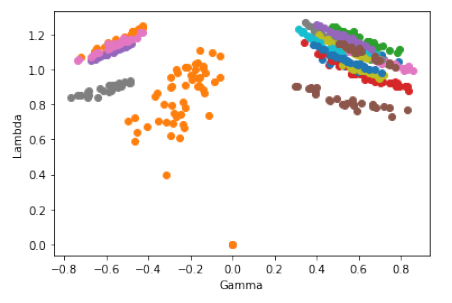}
	\caption{Learnt STP parameters for STPN in ART. Each dot represents one synapse, each color represents the synapses of a single neuron. There is clearly some clustering of the parameters for each neuron at a specific sign for $\gamma$, but with different learning rates for different synapses.}
	\label{fig:fireworks_art}
\end{figure}


\end{document}